\documentclass[preprint,3p,number,nopreprintline]{elsarticle}

\usepackage{amsmath,amssymb,amsfonts,amsthm,mathtools,bm}
\usepackage{algorithm}
\usepackage[noend]{algpseudocode}
\usepackage{booktabs}
\usepackage{float}
\usepackage{subcaption}
\usepackage{placeins}
\usepackage{xcolor}
\usepackage{microtype}
\usepackage{enumitem}
\usepackage[colorlinks=true,allcolors=blue]{hyperref}
\usepackage[nameinlink,noabbrev,capitalise]{cleveref}
\usepackage{etoolbox}
\makeatletter
\patchcmd{\fnmark}{\unskip\comma\fnotenum}
  {\unskip\comma\ifcase\fnotenum\or$\ast$\or$\dagger$\fi}
  {}{\PackageError{authornotes}{Cannot patch author marks}{}}
\patchcmd{\fntext}{\def\thefootnote{\thefnote}}
  {\def\thefootnote{\ifcase\c@fnote\or$\ast$\or$\dagger$\fi}}
  {}{\PackageError{authornotes}{Cannot patch author footnotes}{}}
\makeatother
\usepackage{tikz}
\usetikzlibrary{positioning,arrows.meta,matrix}

\newtheorem{theorem}{Theorem}
\newtheorem{proposition}[theorem]{Proposition}

\newtheorem{corollary}[theorem]{Corollary}
\newtheorem{assumption}[theorem]{Assumption}
\theoremstyle{remark}

\newcommand{\R}{\mathbb{R}}
\newcommand{\E}{\mathbb{E}}

\newcommand{\pss}{p_{\mathrm{ss}}}
\newcommand{\qth}{q_{\theta}}
\newcommand{\Uth}{U_{\theta}}
\newcommand{\Lepr}{\mathcal{L}_{\mathrm{EPR}}}
\newcommand{\Lhjb}{\mathcal{L}_{\mathrm{HJB}}}
\newcommand{\withsd}[2]{\ensuremath{#1{\scriptstyle\pm#2}}}

\crefname{assumption}{Assumption}{Assumptions}
\Crefname{assumption}{Assumption}{Assumptions}

\journal{Journal of Computational Physics}

\begin{document}
\begin{frontmatter}

\title{EPR-NF: Simulation-Free Learning of Non-Equilibrium
Energy Landscapes with Self-Bootstrapping Normalizing Flows}

\author[nus]{Yue Zhao\fnref{equal}}
\author[cds]{Zichen Liu\fnref{equal}}
\author[lmam]{Jin Lin}
\author[cds,lmam,cmlr]{Tiejun Li\fnref{corresponding}}

\address[nus]{Institute for Functional Intelligent Materials, National University of Singapore.}
\address[cds]{Center for Data Science, Peking University.}
\address[lmam]{LMAM and School of Mathematical Sciences, Peking University.}
\address[cmlr]{Center for Machine Learning Research, Peking University.}

\fntext[equal]{Equal contribution.}
\fntext[corresponding]{Corresponding author. Email address: \href{mailto:tieli@pku.edu.cn}{\texttt{tieli@pku.edu.cn}}}

\begin{abstract}
The non-equilibrium landscape $U=-D\log\pss$ provides a global characterization and a stability measure of a stochastic dynamical system. However, numerical construction of the landscape itself is challenging for high dimensional systems. Based on previously designed EPR-Net formulation, which learns the landscape by a force projection strategy through the entropy production
rate (EPR) variational principle, we introduce EPR-NF, a self-bootstrapping normalizing-flow
method that learns from a prescribed drift without precomputed stationary
trajectories. In EPR-NF, a single normalized density model represents the landscape,
generates training samples, and provides the spatial derivatives required by
the governing equation. Better accuracy is achieved by using the enhanced EPR, which combines a
self-sampled EPR loss with the viscous Hamilton--Jacobi--Bellman (HJB)
residual of the stationary Fokker--Planck equation evaluated at enhanced points covering poorly sampled regions. For pure self-bootstrap EPR, we prove local convergence of the population
iteration to the true stationary distribution under mild conditions. Experiments cover multistable and oscillatory systems, including a twelve-dimensional Gaussian mixture and full-dimensional models of
three- and fifty-two-dimensional biological networks. We also present strategies to learn reduced landscapes through a reduced-force learning. Neither full-dimensional training nor dimensionality reduction requires simulated stationary samples. These results support normalized, self-sampling PDE solvers as an alternative to trajectory-based landscape reconstruction.
\end{abstract}


\end{frontmatter}

\section{Introduction}
\label{sec:introduction}

\begin{figure}[t]
\centering
\includegraphics[width=\linewidth]{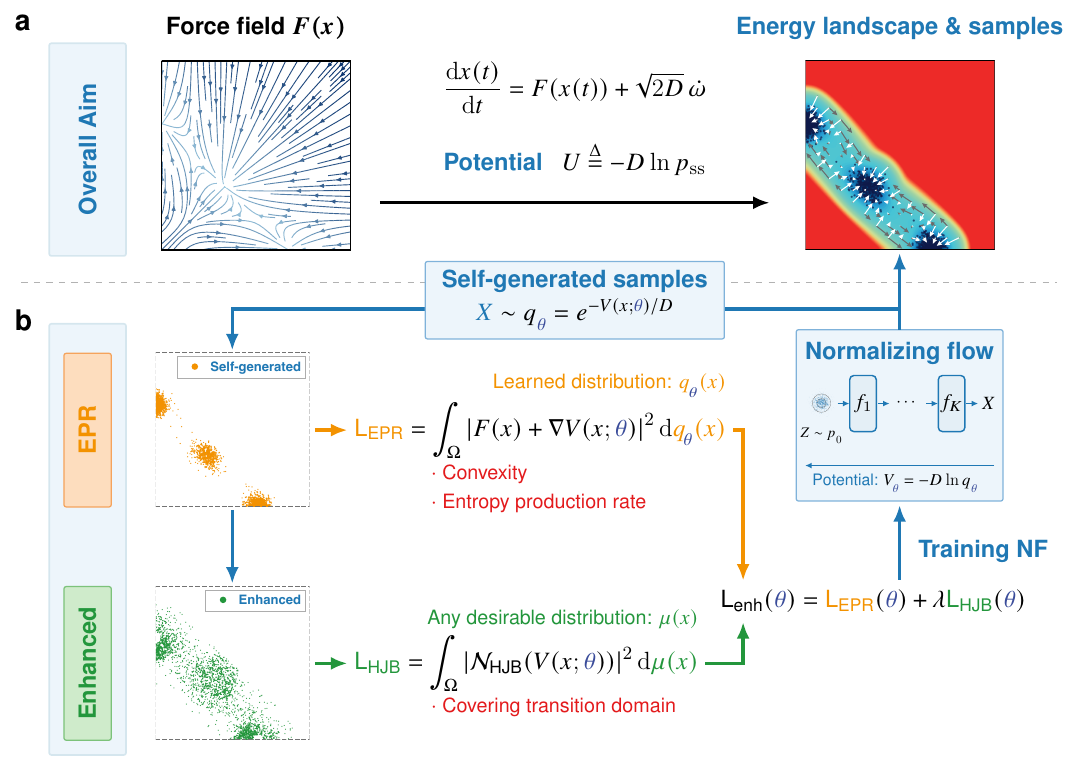}
\caption{Enhanced EPR illustrated with the learned double-well example. (a) Given the drift and diffusion, one normalizing flow supplies a density, its potential landscape, and direct samples. (b) Unperturbed flow samples define the EPR force projection; noise or biased dynamics expand their coverage to form HJB collocation points. Both losses update the same model, whose refreshed samples close the self-bootstrap loop. The inset plots use saved data from the final model on $[0,3]^2$ at $D=0.05$; orange and green points are model-generated and enhanced samples, respectively. The landscape panel also shows computed gradient and nongradient forces in the low-potential region.}
\label{fig:schematic}
\end{figure}

Energy landscapes provide a geometric description of stochastic systems
with competing states. Originating in Waddington's picture of cell-fate
decisions, they have become quantitative tools for biochemical networks,
development, and driven systems \citep{Waddington57,Wang15Review,Fang19Review}.
Basins represent frequently occupied or metastable regimes, their relative
depths summarize stationary populations, and ridges and saddles organize
noise-induced switching. This scalar representation connects stochastic
dynamics to questions of stability, state selection, and transition
structure. Applications include cell-cycle robustness \citep{Wang10},
genetic switching \citep{Lv14}, epithelial--mesenchymal transitions
\citep{Li16}, and cell differentiation with proliferation and death
\citep{Shi22}.

We consider the overdamped stochastic differential equations (SDEs)
\begin{equation}
  \mathrm{d}X_t=F(X_t)\,\mathrm{d}t+\sqrt{2D}\,\mathrm{d}W_t,
  \qquad X_t\in\Omega\subseteq\R^d,
  \label{eq:sde}
\end{equation}
with suitable initial and boundary value conditions (see detailed setup in \cref{sec:preliminaries}), 
where $F$ is the drift, $D>0$ is a constant isotropic diffusion coefficient,
and $W_t$ is the standard Brownian motion. Under appropriate ergodicity
assumptions, the system has a unique stationary density $\pss$. Following
EPR-Net \citep{zhao2024nsr}, we study the finite-noise landscape
\begin{equation}
  U(x)=-D\log\pss(x)+C,
  \label{eq:landscape}
\end{equation}
with an immaterial additive constant $C$. This landscape provides a stability measure of the system states at a given noise level, which is also related to the global quasi-potential
characterizing small-noise asymptotics through an action principle or a
first-order Hamilton--Jacobi equation \citep{Zhou16,Cameron12,Ge12,Zhou12}.
First-exit times and likely transition paths provide complementary
kinetic descriptions of noise-induced escape and epigenetic switching
\citep{Maier93,Aurell02}. At fixed, nonzero diffusion, the finite-noise
landscape retains the relative probability mass of competing states and
provides a global picture of multistable and oscillatory stationary
structures. This is the object reconstructed throughout this work.

At equilibrium, detailed balance removes the stationary probability
current and the drift is a gradient field. Non-equilibrium systems sustain
probability currents through external driving or chemical energy input
\citep{Fang19Review,Zhang12}. Their landscape is formed by the global
transport of probability current, so the landscape reconstruction requires solving for the
stationary distribution. Direct grid discretization of the stationary
Fokker--Planck equation is generally infeasible, which requires $m^d$ unknowns for $m$ nodes per coordinate
in $d$ dimensions, making high-dimensional biological networks particularly
challenging. The computational task involves resolving both high-probability
basins and the sparsely occupied regions separating them. Accurate basin
weights describe state selection, while the geometry between basins
provides information about barriers. Both features are needed to assess
the quality of a learned landscape.

Neural methods reduce dependence on explicit grids by learning
quasi-potentials from observed dynamics \citep{Lin21} or finite-noise
invariant densities from trajectories \citep{Lin22,Lin23}, including
transient observations \citep{Lin23}. EPR-Net introduces a new variational
formulation: the negative landscape gradient is the
stationary-density-weighted orthogonal projection of the drift onto
gradient fields \citep{zhao2024nsr}. The resulting entropy-production-rate
(EPR) functional is quadratic and convex in the scalar potential, which ensures the robustness and efficiency of the computation. However, its evaluation requires precomputed stationary samples. In metastable systems, trajectories may remain within one basin for long time; collecting enough transitions to estimate relative basin populations can cost much computational effort. The scalar potential also leaves density normalization and direct sample generation as separate tasks. Overcoming these issues is the central aim of this paper.

Physics-informed neural networks enforce governing equations through
residual minimization \citep{Raissi19}. Normalizing flows provide normalized
densities and direct samples through invertible transformations
\citep{rezende2015variational,papamakarios2021normalizing}, enabling a model
to supply its own collocation points. Tang et al.~\citep{tang2022adda}
introduced ADDA-KR, which combines a KRnet density with Fokker--Planck
residual minimization and adaptive sampling from the current model.
Zeng et al.~\citep{zeng2025boundedkrnet} developed B-KRnet for bounded
domains and extended the construction to mixed bounded--unbounded domains.
These approaches connect density representation and adaptive PDE solution.
For EPR learning, the corresponding challenge is to replace the unknown
stationary sampling measure while retaining the force-projection structure.
The sampling distribution evolves with the learned density, creating a
feedback between where the equation is enforced and how the model is
updated. Effective training must therefore maintain broad coverage as
probability concentrates near the attractors.

We introduce \emph{EPR-NF}, a self-bootstrapping normalizing-flow method
that couples EPR force projection with the viscous
Hamilton--Jacobi--Bellman (HJB) form of the stationary Fokker--Planck
equation. We call this training formulation \emph{Enhanced EPR}. Our idea to overcome the stationary sampling is by iteration through EPR variational formulation. We utilize a single
flow $q_\theta$ to represent the landscape
$U_\theta=-D\log q_\theta$, generate training samples, and supply the
spatial derivatives required by both losses. EPR uses samples from a
frozen copy of the current model, preserving weighted force projection
during each update. Enhanced HJB collocation extends the stationary
constraint to poorly sampled regions, while refreshed model samples
close the self-bootstrap loop. A diffusion continuation begins at higher
noise to promote exploration and then resolves the landscape at the
target diffusion. The final converged flow through self-bootstrapping provides both stationary sampler and non-equilibrium potential. The learned full density also supplies samples for
reduced landscape construction in selected coordinates, connecting to
reaction-coordinate and free-energy methods \citep{Zhang16,liu2026fesfm}.
Here, simulation-free means that every EPR-NF training stage, including
dimensionality reduction, operates without simulated stationary samples.

Our main contributions are:
\begin{itemize}[leftmargin=*,itemsep=2pt]
  \item We develop EPR-NF, a normalized self-sampling landscape solver that combines EPR force projection with enhanced HJB collocation in one flow. The model supplies its own training samples, landscape, and spatial derivatives; diffusion continuation and enhanced collocation extend coverage during training.
  \item For the pure self-bootstrap EPR population iteration, we identify the stationary density as the unique fixed-point density and establish local geometric convergence under an explicit stability condition. An inexact-update bound further quantifies the effect of approximate inner solves.
  \item We evaluate multistable and oscillatory systems up to $52$ dimensions without stationary training samples, including model-based reduced landscape construction. The learned landscapes recover the principal structures, and Enhanced EPR achieves lower landscape errors than the stationary-data KL baseline on all three two-dimensional benchmarks.
\end{itemize}

This paper is organized as follows. Section~\ref{sec:preliminaries} introduces the stationary problem and EPR-Net. Section~\ref{sec:method} presents EPR-NF and model-based reduction, and Section~\ref{sec:theory} analyzes pure self-bootstrap EPR. Section~\ref{sec:experiments} reports the numerical results, followed by discussion and conclusions. Proofs, additional experiments, and implementation details are collected in the appendices.

\section{Mathematical Setup and EPR-Net}
\label{sec:preliminaries}

Consider the SDEs \eqref{eq:sde} with isotropic diffusion constant $D$. Its probability density function $p(x,t)$ satisfies the Fokker-Planck equation
\begin{equation}
  \partial_t p=-\nabla\cdot(Fp)+D\Delta p.
  \label{eq:fpe-time}
\end{equation}
The steady state distribution density $\pss$ satisfies
\begin{equation}
  -\nabla\cdot(F\pss)+D\Delta\pss=0 \quad\text{in }\Omega.
  \label{eq:stationary-fpe}
\end{equation}
When the domain $\Omega=\R^d$, we impose the boundary condition
\begin{equation}
\pss(x)\rightarrow 0\quad \text{sufficiently fast as}\ |x|\rightarrow\infty;
\end{equation}
on a bounded domain we impose the reflecting boundary condition
\begin{equation}
  J_{\mathrm{ss}}\cdot n=0 \quad\text{on }\partial\Omega,
  \label{eq:no-flux}
\end{equation}
where $J_{\mathrm{ss}}=F\pss-D\nabla\pss$ is the steady state probability current.   Using
\eqref{eq:landscape}, the current and current velocity are
\begin{equation}
  J_{\mathrm{ss}}=\pss(F+\nabla U),
  \qquad
  v=\frac{J_{\mathrm{ss}}}{\pss}=F+\nabla U,
  \label{eq:current-velocity}
\end{equation}
where $U(x)=-D\log \pss(x)$ is the target non-equilibrium potential for the considered system. We have 
$C=0$ in \eqref{eq:landscape} for the normalized density $\pss$. The system satisfies detailed balance condition precisely when $J_{\mathrm{ss}}=0$, i.e. $F=-\nabla U$.

\subsection{The EPR variational principle}

EPR-Net constructs the landscape by minimizing a force-projection loss on
the stationary density \citep{zhao2024nsr}.  For an admissible scalar function $V$, the EPR loss is
\begin{equation}
  \Lepr(V;\pss)
  =\E_{X\sim\pss}\|F(X)+\nabla V(X)\|^2.
  \label{eq:epr-loss}
\end{equation}
The projection identity established in \citep{zhao2024nsr} decomposes it as
\begin{equation}
  \Lepr(V;\pss)
  =\Lepr(U;\pss)
  +\E_{\pss}\|\nabla(V-U)\|^2.
  \label{eq:pythagorean}
\end{equation}
Thus $U$ is the unique minimizer up to an additive constant, and the
minimum is proportional to the steady entropy production rate in non-equilibrium statistical physics \citep{Fang19Review,Zhang12},
\begin{equation}
  \Lepr(U;\pss)=\E_{\pss}\|v\|^2=D e_p,
  \qquad
  e_p=D^{-1}\E_{\pss}\|v\|^2.
  \label{eq:epr-minimum}
\end{equation}
The expectation in \eqref{eq:epr-loss} is evaluated on stationary samples,
so applying this formulation directly requires representative stationary
samples.

\subsection{HJB enhancement}

The stationary Fokker--Planck equation also yields a pointwise equation for
the landscape.  Substituting $p=Z^{-1}e^{-U/D}$ into
\eqref{eq:stationary-fpe} and dividing by $p>0$ gives
\begin{equation}
  \mathcal{N}_{F,D}(U)
  :=F\cdot\nabla U+\|\nabla U\|^2
    -D\Delta U-D\nabla\cdot F=0.
  \label{eq:hjb}
\end{equation}
This is the viscous stationary HJB form used to enhance EPR learning in the
small-noise regime \citep{zhao2024nsr}.  For a collocation distribution
$\mu$, define
\begin{equation}
  \Lhjb(V;\mu)
  =\E_{X\sim\mu}\left|\mathcal{N}_{F,D}(V)(X)\right|^2.
  \label{eq:hjb-loss}
\end{equation}
The pointwise residual $\mathcal{N}_{F,D}(V)(x)$ can be evaluated at
collocation points chosen to cover regions of interest, especially the important transition regions which are sparsely covered by stationary samples.

\subsection{Reduced force for prescribed coordinates}
\label{sec:reduced-force}

The high dimensional landscape needs to be visualized in a low dimensional space. We perform landscape dimensionality-reduction in prescribed coordinates through a reduced force formulation.  Split $x=(w,y)$, where $y\in\R^m$ are the retained coordinates and $w$ collects
the remaining coordinates, and write $F=(F_w,F_y)$. The reduced landscape in $y$ space can be naturally defined through the marginal stationary density \citep{zhao2024nsr}
\begin{equation}
  \tilde p_{\mathrm{ss}}(y)=\int p_{\mathrm{ss}}(w,y)\,\mathrm{d}w,
  \qquad
  \tilde U(y)=-D\log\tilde p_{\mathrm{ss}}(y).
  \label{eq:reduced-landscape}
\end{equation}
An effective drift compatible with this stationary
marginal is the conditional expectation of the retained force components,
\begin{equation}
  F_{\mathrm{red}}(y)
  =\int F_y(w,y)\,p_{\mathrm{ss}}(w\mid y)\,\mathrm{d}w,
  \qquad
  p_{\mathrm{ss}}(w\mid y)
  =\frac{p_{\mathrm{ss}}(w,y)}{\tilde p_{\mathrm{ss}}(y)}.
  \label{eq:reduced-force}
\end{equation}
This function minimizes the force-regression loss
\begin{equation}
  \mathcal L_{\mathrm{force}}(g)
  =\E_{(W,Y)\sim p_{\mathrm{ss}}}
   \|F_y(W,Y)-g(Y)\|^2.
  \label{eq:force-loss}
\end{equation}
With $F_{\mathrm{red}}$ in place of $F$, the reduced landscape satisfies
the same stationary Fokker--Planck equation and the same viscous HJB
equation \eqref{eq:hjb} in the $y$-coordinates.  The EPR projection and
the HJB residual above therefore apply to $(F_{\mathrm{red}},D)$ without
change, assuming the boundary terms vanish when marginalizing over $w$.
This construction characterizes the stationary marginal; it does not imply
that the projected transient dynamics is Markovian. Equations
\eqref{eq:reduced-force}--\eqref{eq:force-loss} characterize the exact
stationary marginal; our implementation replaces the unknown sampling
density by a learned full-dimensional flow, as described next.

\section{Self-Bootstrapping EPR with Normalizing Flows}
\label{sec:method}

This section presents our EPR-NF method for learning the stationary
landscape from a given drift and diffusion. We first introduce its
self-bootstrapping principle, then describe the flow representation,
training objective, enhanced sampling, and model-based reduction.

Starting from a current density $q_k$, the idealized EPR step holds $q_k$ fixed and updates the landscape and density according to
\begin{equation}
 V_{k+1}\in\arg\min_V\mathbb E_{X\sim q_k}\|F(X)+\nabla V(X)\|^2,
 \qquad q_{k+1}=\frac{e^{-V_{k+1}/D}}{\int_\Omega e^{-V_{k+1}/D}\,\mathrm{d}x}.
\end{equation}
Thus the current density supplies samples for the next force projection, and the updated density in turn supplies samples for the following step. In practice, a single flow parameterizes the density and receives gradient updates using frozen model samples, enhanced HJB collocation, and diffusion continuation. The exact population iteration is analyzed in Section~\ref{sec:theory}; the following subsections describe the practical objective, sampling mechanisms, and reduction procedure.

\subsection{Normalized landscape parameterization}

Let $Z\sim p_0$ be a tractable base distribution and let
$X=T_\theta(Z)$, where $T_\theta:\R^d\to\Omega$ is invertible.  The induced
density is
\begin{equation}
  \log \qth(x)
  =\log p_0(T_\theta^{-1}(x))
   +\log\left|\det\nabla_x T_\theta^{-1}(x)\right|.
  \label{eq:change-of-variables}
\end{equation}
We define the model landscape as
\begin{equation}
  \Uth(x)=-D\log\qth(x).
  \label{eq:flow-landscape}
\end{equation}
It follows that
\begin{equation}
  \nabla\Uth=-D\nabla\log\qth,
  \qquad
  \Delta\Uth=-D\Delta\log\qth.
  \label{eq:score-laplacian}
\end{equation}
Thus the force-projection term, the HJB residual, normalized likelihoods, and
samples are all obtained from the same model.

Because \eqref{eq:hjb} contains a Laplacian, the flow must support numerically stable evaluation of
second spatial derivatives.  We therefore use a Bernstein autoregressive flow and
incorporate the two implementation modifications described in \ref{app:bnf}.  Other flow families
may be substituted if they provide an accurate log density and sufficiently
regular second derivatives.

\subsection{Objective}
\label{sec:objective}

The Enhanced EPR objective combines force projection and a stationary PDE constraint:
\begin{equation}
  \mathcal{J}_k(\theta)=\Lepr+\lambda\Lhjb.
  \label{eq:objective}
\end{equation}
The on-policy EPR projection and the pointwise stationary constraint are
\begin{equation}
  \Lepr=\E_{X\sim q_{\bar\theta_k}}
    \|F(X)-D\nabla\log\qth(X)\|^2,
    \label{eq:objective-epr}
\end{equation}
\begin{equation}
  \Lhjb=\E_{X\sim\mu_k}
    \left|\mathcal{N}_{F,D}(\Uth)(X)\right|^2.
    \label{eq:objective-hjb}
\end{equation}
Neither term requires observations of the target density.

The two sampling measures serve different purposes. For EPR, samples are
drawn from a frozen copy of the flow,
$\bar\theta_k=\operatorname{stopgrad}(\theta_k)$, and refreshed periodically.
The sampling locations are held fixed during differentiation, so each update
retains the weighted force-projection interpretation under the current
sampling measure. When this measure is the stationary density, the population
projection recovers the stationary landscape; self-bootstrapping updates the
weight using the learned flow. Differentiating through the sampling density
would introduce an additional density-variation term and define a different
objective. Between refreshes,
the sampling density may lag behind the current model; this lag is accounted
for in \cref{sec:theory}. The HJB measure $\mu_k$ is constructed using the
enhancement mechanisms in \cref{sec:enhancement} to increase coverage of
barriers and other low-density regions. These enhanced points enter only the
HJB loss; the EPR loss uses unperturbed flow samples. The HJB measure is also frozen within each
update. The coefficient $\lambda>0$ sets the strength of the HJB constraint relative to EPR;
single-term controls and numerical choices are reported in
\ref{app:experimental-details}.

The HJB residual can be written directly in terms of the model score
$s_\theta=\nabla\log\qth$:
\begin{equation}
  \mathcal{N}_{F,D}(\Uth)
  =-D F\cdot s_\theta+D^2\|s_\theta\|^2
    +D^2\nabla\cdot s_\theta-D\nabla\cdot F.
  \label{eq:hjb-score}
\end{equation}
Evaluating this residual requires consistent first and second spatial
derivatives of the log density. The Bernstein-flow implementation and its
derivative calculations are described in \ref{app:bnf}.

\subsection{Diffusion continuation and training procedure}
\label{sec:schedules}

To improve exploration in metastable systems, training begins at an elevated
diffusion and gradually approaches the target $D$. This continuation promotes
broad exploration at higher diffusion before resolving the
small-noise landscape at the target diffusion.
At each stage, the same diffusion $D_k$ is used in the landscape
$-D_k\log q_\theta$, both losses, and the collocation procedure. Training
then continues at the target diffusion. The schedule and optimizer settings
are specified in \ref{app:experimental-details}.
\Cref{alg:epr-nf,fig:schematic} summarize the resulting learning loop.

\begin{algorithm}[t]
\caption{Enhanced EPR training in EPR-NF}
\label{alg:epr-nf}
\begin{algorithmic}[1]
\Require drift $F$, target diffusion $D$, HJB weight $\lambda$,
         diffusion schedule $D_k\to D$, redraw interval $R$
\State initialize $q_\theta$ near uniform on its support, and an empty buffer
       for the HJB measure $\mu_k$
\For{$k=0,1,\ldots,K-1$}
  \State set the current diffusion $D_k$
  \If{$k\equiv 0 \pmod{R}$}
    \State set $\bar\theta\gets\operatorname{stopgrad}(\theta)$ and draw a
           fresh EPR pool from $q_{\bar\theta}$
  \EndIf
  \State refresh $\mu_k$: perturb pool samples by isotropic Gaussian noise,
         or advance walkers by \eqref{eq:biased-dynamics}
  \For{each minibatch}
    \State draw $X$ from the frozen EPR pool
    \State evaluate $F(X)$ and $s_\theta(X)=\nabla\log q_\theta(X)$, and form
           $\Lepr$ by \eqref{eq:objective-epr} at diffusion $D_k$
    \State draw $Y\sim\mu_k$
    \State evaluate $s_\theta(Y)$, form $\mathcal{N}_{F,D_k}(U_\theta)(Y)$ by
           \eqref{eq:hjb-score}, and form $\Lhjb$ by \eqref{eq:objective-hjb}
    \State set $\mathcal{J}_k(\theta)=\Lepr+\lambda\Lhjb$ as in
           \eqref{eq:objective}, and take one Adam step on $\theta$
  \EndFor
\EndFor
\State \Return $q_\theta$, the landscape $U_\theta=-D\log q_\theta$ of
       \eqref{eq:flow-landscape}, and the sampler $T_\theta$ of
       \eqref{eq:change-of-variables}
\end{algorithmic}
\end{algorithm}

\subsection{Enhanced HJB collocation}
\label{sec:enhancement}

Samples from either $\pss$ or a well-trained $q_\theta$ rarely visit the
low-density regions between metastable states, although these regions
contain important barrier geometry. We enrich HJB collocation in these
regions to evaluate the pointwise stationary residual. Two mechanisms are
used, alone or in combination; none of them simulates the target SDE
\eqref{eq:sde}, and enhanced points are excluded from the EPR loss.

\paragraph{Noise inflation}
HJB points are on-policy samples perturbed by isotropic Gaussian noise,
$x+\sigma\xi$, $\xi\sim N(0,I)$, with out-of-domain points reflected into
$\Omega$; $\sigma=0$ recovers unmodified flow samples.  This mechanism has the lowest computational cost of the two and suffices
when the required spread is a fraction of the domain width.

\paragraph{Biased dynamics with energy-window filtering}
Starting from on-policy samples, walkers are advanced under
\begin{equation}
  \mathrm{d}X_t=\bigl[F(X_t)-s\,D_k\nabla\log\qth(X_t)\bigr]\,\mathrm{d}t+\sqrt{2\eta D_k}\,\mathrm{d}W_t,
  \label{eq:biased-dynamics}
\end{equation}
which uses only drift and score evaluations. The added score term opposes
the model's gradient attraction. If the model landscape is exact, $s=1$
leaves only the stationary current velocity in the drift, while $s>1$ adds
an uphill component. To select collocation points, we set
$U_k(x)=-D_k\log q_\theta(x)$ and take $U_{\mathrm{ref}}$ as the minimum
of $U_k$ over a fresh batch of flow samples. A candidate $x$ is retained
when $aD_k\le U_k(x)-U_{\mathrm{ref}}\le bD_k$, where $aD_k$ and $bD_k$
are the lower and upper bounds of the energy window. The lower bound reduces the
number of points near the sampled potential minimum, while the upper bound
limits the range of high-energy points. This selection directs HJB
evaluations toward regions that receive less stationary mass. The reference
is a sample minimum, and the window and walker settings are specified in
\ref{app:experimental-details}.

\Cref{fig:enhancement} illustrates the two mechanisms on the learned
double-well landscape.

\subsection{Full-dimensional learning and model-based reduction}
\label{sec:model-reduction}

For a high-dimensional system, we train $q_\theta(w,y)$ directly from the
full drift $F$ and diffusion $D$, using the self-bootstrap procedure above.
The trained flow then supplies samples $X=(W,Y)\sim q_\theta$ for
conditional-drift regression. We freeze this full-dimensional model
before learning the conditional drift and reduced landscape.

\paragraph{Reduced-force learning}
Following EPR-Net \citep{zhao2024nsr}, we use dimension reduction (DR) to refer to the reduced-force learning route. Here the conditional drift is learned from full-flow samples. We fit a vector field $g_\eta$ using model samples and the known drift,
\begin{equation}
 \mathcal L_{\mathrm{force}}^{q_\theta}(g_\eta)
 =\mathbb E_{(W,Y)\sim q_\theta}\|F_y(W,Y)-g_\eta(Y)\|^2.
 \label{eq:model-force-loss}
\end{equation}
Its population minimizer is $g_q(y)=\mathbb E_{q_\theta}[F_y(W,Y)\mid Y=y]$.
Keeping $g_\eta$ fixed, we train a second, lower-dimensional normalized flow
$r_\phi(y)$ with self-sampled EPR and enhanced HJB losses for $(g_\eta,D)$.
This produces $V_\phi=-D\log r_\phi$ and a direct reduced sampler. No
stationary SDE samples enter either regression or landscape training.

DR solves the stationary problem defined by the learned conditional drift.
When $q_\theta=p_{\mathrm{ss}}$, the regression and reduced solve are exact,
and the boundary assumptions of \cref{sec:reduced-force} hold, it recovers
the true stationary marginal.
The same model-based sampling strategy applies to both dimensionality-reduction
methods in EPR-Net \citep{zhao2024nsr}: reduced-force learning and gradient
projection (GP).

\section{Theoretical Analysis}
\label{sec:theory}

We analyze the population self-bootstrap iteration driven by the EPR loss
alone, at a fixed diffusion $D>0$. The result identifies its unique
self-consistent density and proves local geometric convergence under an
explicit stability condition. All proofs are collected
in \ref{app:proofs}.

\subsection{The pure EPR population update}

The following assumption specifies the stationary problem and the function
space used to analyze the population iteration.

\begin{assumption}
\label{ass:regularity}
\label{ass:bootstrap}
For some $0<\alpha<1$, $\Omega$ is a bounded connected $C^{2,\alpha}$ domain,
$F\in C^{1,\alpha}(\overline\Omega)$, and the reflecting stationary
Fokker--Planck problem has a unique normalized density
$p=\pss\in C^{2,\alpha}(\overline\Omega)$ with $\min p>0$.
Set $U=-D\log p$, $\bar U=U-\int_\Omega pU\,\mathrm{d}x$, and
\[
 \mathcal X=\{h\in C^{2,\alpha}(\overline\Omega):
       \partial_nh=0,\ \int_\Omega ph\,\mathrm{d}x=0\},
 \qquad \|h\|_{\mathcal X}=\|h\|_{C^{2,\alpha}}.
\]
The input potentials lie in $\bar U+\mathcal X$. The inner optimization
allows all $H^1$ potential variations, modulo additive constants.
\end{assumption}

For a frozen positive density $r$, define
\begin{equation}
 \mathcal L(V;r)=\int_\Omega r\|F+\nabla V\|^2\,\mathrm{d}x,
 \qquad q_W=\frac{e^{-W/D}}{\int_\Omega e^{-W/D}\,\mathrm{d}x}.
 \label{eq:population-epr}
\end{equation}
The potential and density updates are
\begin{equation}
 \mathcal S(W)=\underset{V\in H^1(\Omega),\ \int pV=0}{\arg\min}
                      \mathcal L(V;q_W),
 \qquad W_{k+1}=\mathcal S(W_k),\quad q_{k+1}=q_{W_{k+1}}.
 \label{eq:self-map}
\end{equation}
The weight $q_W$ is held fixed during minimization. Strict convexity in
$\nabla V$ and the gauge give a unique minimizer. Its natural boundary
condition and Euler--Lagrange equation are
\begin{equation}
 \nabla\cdot\bigl(q_W(F+\nabla\mathcal S(W))\bigr)=0,
 \qquad (F+\nabla\mathcal S(W))\cdot n=0.
 \label{eq:weighted-el}
\end{equation}
Thus the theorem concerns repeated frozen-measure EPR minimization, rather
than differentiating through the sampling density.

\subsection{The stationary density is the unique fixed point}

\begin{proposition}
\label{prop:fixed-point}
\label{prop:consistency}
Under \cref{ass:bootstrap}, $\mathcal S(\bar U)=\bar U$. More precisely,
\begin{equation}
 \mathcal L(V;p)-\mathcal L(\bar U;p)
       =\int_\Omega p\|\nabla(V-\bar U)\|^2\,\mathrm{d}x.
 \label{eq:epr-projection-bootstrap}
\end{equation}
\end{proposition}

\begin{theorem}
\label{thm:selfconsistent}
If an admissible $V$ is stationary for $\mathcal L(\cdot;q_V)$ against
all $H^1$ variations, with $q_V$ frozen, then $q_V=p$. Consequently $p$
is the unique fixed-point density of the pure EPR population update.
\end{theorem}

Indeed, self-consistency turns \eqref{eq:weighted-el} into
$\nabla\cdot(q_VF-D\nabla q_V)=0$, with reflecting boundary.
This also identifies the density limit of any convergent exact iteration.
The next result establishes attraction to this fixed point.

\subsection{Local convergence of self-bootstrap EPR}

Let $v=F+\nabla U$ be the stationary current velocity; it satisfies
$\nabla\cdot(pv)=0$ and $v\cdot n=0$. Define the weighted Neumann operator
\begin{equation}
 \mathcal A h=-p^{-1}\nabla\cdot(p\nabla h),\qquad
 P_pf=f-\int_\Omega pf\,\mathrm{d}x,\qquad
 \mathcal R=\mathcal A^{-1}P_p.
 \label{eq:epr-neumann}
\end{equation}
Here $\mathcal A^{-1}$ maps $p$-mean-zero $C^{0,\alpha}$ functions into
$\mathcal X$; its boundedness is established in \ref{app:proofs}.
The linearized EPR update is
\begin{equation}
 \mathcal B h=-\frac1D\mathcal A^{-1}(v\cdot\nabla h),
 \qquad b=\|\mathcal B\|_{\mathcal X\to\mathcal X}.
 \label{eq:epr-linearization}
\end{equation}
The inverse is applicable because $\int p\,v\cdot\nabla h=0$.
Let $C_{\mathrm{H}}$ be a H\"older product constant satisfying
$\|\nabla h\cdot\nabla g\|_{C^{0,\alpha}}
 \le C_{\mathrm{H}}\|h\|_{\mathcal X}\|g\|_{\mathcal X}$,
and put $c=\|\mathcal R\|C_{\mathrm{H}}/D$.

\begin{theorem}
\label{thm:bootstrap-convergence}
Under \cref{ass:bootstrap}, assume $b<1$. Choose $b<\kappa<1$ and
$r>0$ such that $cr<1$ and $b/(1-cr)\le\kappa$.
For any $W_0=\bar U+h_0$ with $h_0\in\mathcal X$ and
$\|h_0\|_{\mathcal X}\le r$, every exact update in \eqref{eq:self-map}
is well defined in $\bar U+\mathcal X$ and
\begin{equation}
 \|W_k-\bar U\|_{\mathcal X}
       \le\kappa^k\|W_0-\bar U\|_{\mathcal X}.
 \label{eq:epr-local-rate}
\end{equation}
In particular $q_{W_k}\to p$ and $-D\log q_{W_k}\to U$ in
$C^{2,\alpha}(\overline\Omega)$.
\end{theorem}

The condition $b<1$ can be ensured by a sufficiently small stationary
current relative to diffusion and the Neumann inverse bound:
\begin{equation}
 b\le\frac{C_{\mathcal A}C_{\mathrm{prod}}}{D}
             \|v\|_{C^{0,\alpha}}<1,
 \qquad C_{\mathcal A}=\|\mathcal A^{-1}\|,
 \label{eq:epr-small-current}
\end{equation}
where $C_{\mathrm{prod}}$ bounds
$\|v\cdot\nabla h\|_{C^{0,\alpha}}$ by
$\|v\|_{C^{0,\alpha}}\|h\|_{\mathcal X}$.
This sufficient condition covers non-equilibrium systems with weak
circulation. At detailed balance, $v=0$ and every positive frozen weight
has the same EPR minimizer $U+C$, so an exact inner solve recovers the
stationary density in one step. The proof of the local theorem uses the
exact error identity
\begin{equation}
 \mathcal S(\bar U+h)-\bar U=(I+\mathcal K_h)^{-1}\mathcal B h,
 \qquad \mathcal K_hg=\frac1D\mathcal R(\nabla h\cdot\nabla g),
 \quad \|\mathcal K_h\|\le c\|h\|_{\mathcal X}.
 \label{eq:epr-error-identity}
\end{equation}
Thus convergence follows directly from an initial condition in this
neighborhood.

\subsection{Approximate updates and relation to training}

\begin{corollary}
\label{cor:epr-inexact}
In the ball of \cref{thm:bootstrap-convergence}, consider
$W_{k+1}=\mathcal S(W_k)+e_k$, where $e_k\in\mathcal X$ and
$\|e_k\|_{\mathcal X}\le\eta_k\le(1-\kappa)r$. Then the ball is invariant and
\begin{equation}
 \|W_k-\bar U\|_{\mathcal X}
 \le\kappa^k\|W_0-\bar U\|_{\mathcal X}
       +\sum_{j=0}^{k-1}\kappa^{k-1-j}\eta_j.
 \label{eq:epr-inexact-bound}
\end{equation}
Thus $\eta_k\to0$ implies convergence to $p$; a uniform error bound
$\eta_k\le\eta$ gives
$\limsup_k\|W_k-\bar U\|_{\mathcal X}\le\eta/(1-\kappa)$.
\end{corollary}

The bound provides a criterion for controlling inexact inner solves in
local population EPR iteration: representation, optimization, quadrature,
and sampling-lag errors enter through their combined effect on the update.
The analysis applies on the smooth bounded domain specified above, under
the no-flux boundary condition and the stability condition $b<1$.

\section{Experiments}
\label{sec:experiments}

We evaluate three aspects of EPR-NF: reconstruction of multistable and
oscillatory landscapes, full-dimensional density learning in higher
dimensions, and construction of reduced landscapes from a learned full density
without stationary training samples. The systems follow the EPR-Net benchmark suite
\citep{zhao2024nsr}. All reported results use the final training checkpoint.
Configuration selection, numerical settings,
and supplementary diagnostics are documented in \ref{app:experiments}.

\subsection{Evaluation protocol}
\label{sec:protocol}

The experiments use a Bernstein autoregressive flow. Density evaluation
uses the triangular Jacobian of the forward transformation, while sampling
inverts the transformation coordinate by coordinate using bisection. We
compute the log density, score, and Laplacian together by propagating
second-order derivatives through the flow; direct differentiation provides
a numerical consistency check. Implementation details are given in
\ref{app:bnf}, and numerical settings in \ref{app:experimental-details}.

For the two-dimensional systems and Gaussian mixture, references are
analytic landscapes or numerical stationary Fokker--Planck solutions.
HJB-only controls use the same flow architecture and training budget as
the corresponding Enhanced EPR model. We also include EPR-only training
at moderate noise and a KL baseline that minimizes the forward divergence
$D_{\mathrm{KL}}(p_{\mathrm{ss}}\|q_\theta)$ through maximum-likelihood
training of the same flow on stationary samples. These comparisons distinguish the roles
of the losses and provide a data-based reference for landscape accuracy;
the KL baseline requires a separate sample-generation stage.

The HJB-only control shares the flow-based PDE-solving strategy of ADDA-KR
\citep{tang2022adda} and the PDE-solving framework of B-KRnet
\citep{zeng2025boundedkrnet}: each trains a normalized flow using PDE
residuals and model-based collocation. Here the residual is written for
$U_\theta=-D\log q_\theta$, which rescales the stationary Fokker--Planck
residual by $D/q_\theta$. ADDA-KR minimizes the density-equation residual,
while B-KRnet introduces an auxiliary gradient field for second-order
PDEs.

For low-dimensional systems we report the landscape after shifting its
minimum to zero.  On an evaluation domain $\mathcal D$, the primary errors are
\begin{align}
  \operatorname{rRMSE}
  &=\left(
  \frac{\int_{\mathcal D}|\Uth-U_{\mathrm{ref}}|^2\mathrm{d}x}
       {\int_{\mathcal D}|U_{\mathrm{ref}}|^2\mathrm{d}x}
  \right)^{1/2},
  \label{eq:rmse}
  \\
  \operatorname{rMAE}
  &=\frac{\int_{\mathcal D}|\Uth-U_{\mathrm{ref}}|\mathrm{d}x}
          {\int_{\mathcal D}|U_{\mathrm{ref}}|\mathrm{d}x}.
  \label{eq:metrics}
\end{align}
Modal weights and barrier heights provide complementary measures of
stationary mass and transition geometry.

For the reduced systems, references are empirical landscapes constructed
from projected stationary samples. We report their structural correlation
with the learned landscape and the rRMSE after additive alignment.
These metrics quantify agreement with the empirical stationary marginal.
Training samples for both full-dimensional EPR-NF and the reduction stages
are generated by the model. Reference
construction and evaluation masks
are specified in \ref{app:experimental-details}.
\Cref{tab:low-d-results} summarizes the two-dimensional benchmark results.

\begin{table}[t]
\centering
\caption{Two-dimensional landscape errors (mean $\pm$ standard deviation).
Errors are defined in \eqref{eq:rmse} and~\eqref{eq:metrics}.
HJB-only controls remove the EPR term
from the corresponding Enhanced EPR configuration. KL uses the same flow architecture
with stationary training data. Protocols are given in
\ref{app:experimental-details}.}
\label{tab:low-d-results}
\begin{tabular}{llcc}
\toprule
System & Method & rRMSE & rMAE \\
\midrule
Double well & Enhanced EPR & \withsd{0.0010}{0.0005} & \withsd{0.0009}{0.0006} \\
 & HJB only & \withsd{0.0002}{0.0000} & \withsd{0.0002}{0.0000} \\
 & KL & \withsd{0.2111}{0.0467} & \withsd{0.1395}{0.0274} \\
\addlinespace
Limit cycle & Enhanced EPR & \withsd{0.0943}{0.0027} & \withsd{0.0315}{0.0033} \\
 & HJB only & \withsd{0.5642}{0.0097} & \withsd{0.3608}{0.0224} \\
 & KL & \withsd{0.3570}{0.0122} & \withsd{0.2532}{0.0056} \\
\addlinespace
Multi-stable & Enhanced EPR & \withsd{0.0189}{0.0223} & \withsd{0.0080}{0.0057} \\
 & HJB only & \withsd{0.6876}{0.0170} & \withsd{0.6278}{0.0178} \\
 & KL & \withsd{0.1344}{0.0115} & \withsd{0.0883}{0.0099} \\
\bottomrule
\end{tabular}
\end{table}

\subsection{Two-dimensional benchmark landscapes}

The three two-dimensional systems test basin reconstruction, circulation,
and small-noise multistability within a common evaluation framework. The learned landscapes, model-generated samples, and force decompositions for all three training objectives are compared in \cref{fig:lowd-method-comparison} (\ref{app:lowd-method-comparison}).

\subsubsection{Double-well model}
\label{sec:exp-toy}

The non-equilibrium double well has an analytic stationary landscape and
tests whether self-bootstrapping recovers both basin geometry and modal
mass. At $D=0.05$, Enhanced EPR achieves an rRMSE of $0.0010$ and a modal-weight
$L_1$ error of $0.007$ (\cref{tab:low-d-results};
\cref{fig:landscapes-lowd}(a,d)). The HJB-only control is more accurate on
this example, with an rRMSE of $0.0002$. Both Enhanced EPR and HJB-only training recover the two wells and their connecting landscape; model-generated samples occupy both basins (\cref{fig:lowd-method-comparison}).

At $D=0.1$, EPR-only training recovers the landscape through self-sampled
force projection alone, achieving a mean rRMSE of $0.0060$
and a mean rMAE of $0.0049$. At the same diffusion, we compare with the weak generative sampler (WGS) \citep{cai2026wgs}, which enforces the stationary Fokker--Planck equation in weak form. The tested WGS configurations yield mean rRMSE values of $0.8173$ for the official RealNVP generator and $0.4223$ for the Bernstein-flow variant. The full comparison, including rMAE and learned landscapes, is provided in \ref{sec:exp-wgs}.
The results show that self-sampled force projection can resolve the non-equilibrium landscape without a precomputed stationary sample set. Training diagnostics are provided in \ref{app:experiments}.

\subsubsection{Biochemical limit-cycle model}
\label{sec:exp-limit-cycle}

The second system is a two-variable biochemical network with a stable limit
cycle \citep{Wang08,zhao2024nsr}, at $D=0.1$. Its
stationary landscape forms a curved valley around the cycle, and its drift
has a substantial circulation component.  This example tests whether EPR-NF
can recover a landscape when the force magnitude is dominated by motion
tangent to level sets rather than by the landscape gradient. Enhanced EPR achieves an rRMSE of $0.0943$ and an rMAE of $0.0315$.
The HJB-only control has a substantially larger rRMSE of $0.5642$,
indicating that EPR improves reconstruction under the tested protocol
(\cref{tab:low-d-results}; \cref{fig:landscapes-lowd}(b,e)). The learned potential follows the curved cycle-shaped valley, with direct samples concentrated along it. The force decomposition displays the landscape-gradient component and the circulating component along the valley. The HJB-only landscape shows a broader low-potential region on the same visualization scale (\cref{fig:lowd-method-comparison}). This comparison highlights the value of force projection for resolving landscape geometry in a system with sustained circulation.

\subsubsection{Multi-stable cell-development model}
\label{sec:exp-multistable-2d}

\begin{figure}[t]
\centering
\begin{subfigure}[t]{0.32\textwidth}
\centering
\includegraphics[width=\linewidth]{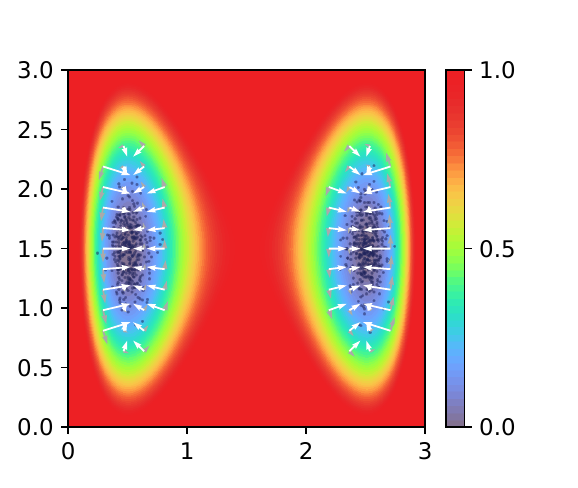}
\caption{Double well: landscape}
\end{subfigure}
\hfill
\begin{subfigure}[t]{0.32\textwidth}
\centering
\includegraphics[width=\linewidth]{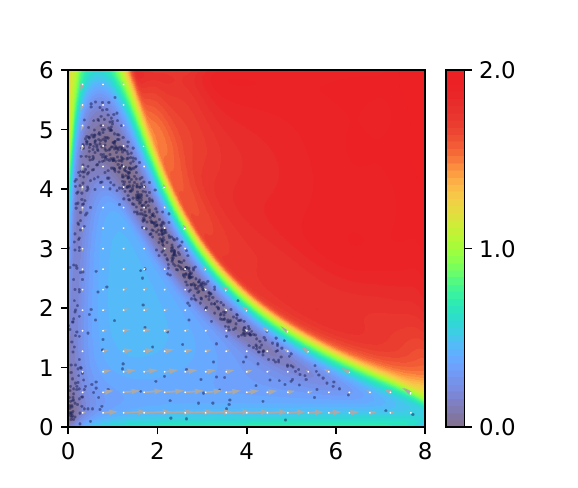}
\caption{Limit cycle: landscape}
\end{subfigure}
\hfill
\begin{subfigure}[t]{0.32\textwidth}
\centering
\includegraphics[width=\linewidth]{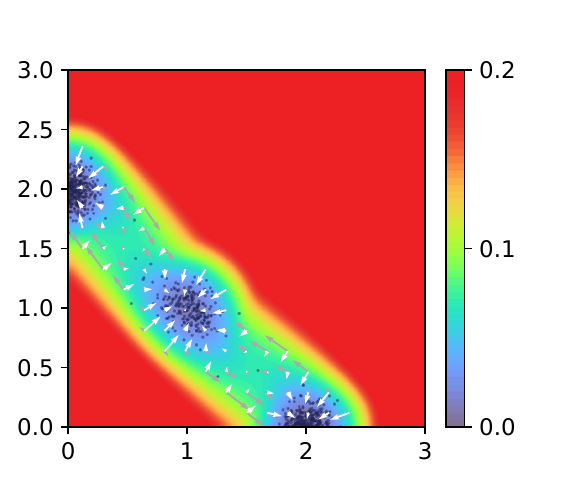}
\caption{Multi-stable: landscape}
\end{subfigure}
\par\medskip
\begin{subfigure}[t]{0.32\textwidth}
\centering
\includegraphics[width=\linewidth]{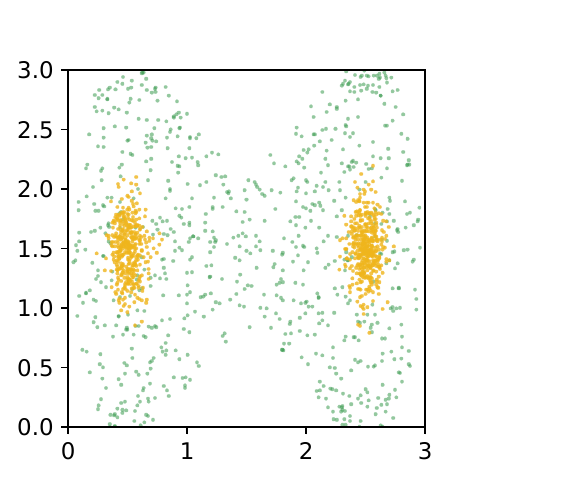}
\caption{Double well: samples}
\end{subfigure}
\hfill
\begin{subfigure}[t]{0.32\textwidth}
\centering
\includegraphics[width=\linewidth]{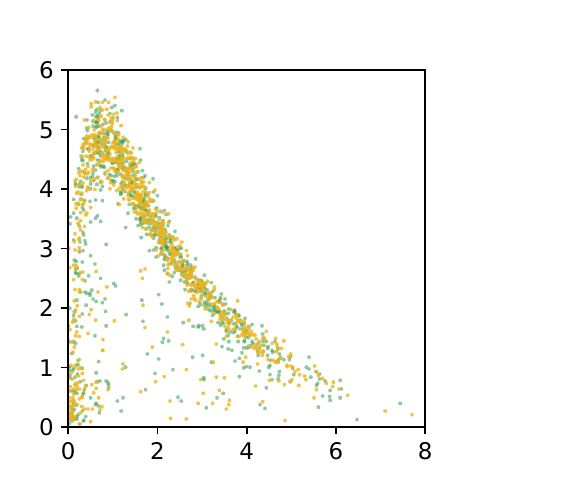}
\caption{Limit cycle: samples}
\end{subfigure}
\hfill
\begin{subfigure}[t]{0.32\textwidth}
\centering
\includegraphics[width=\linewidth]{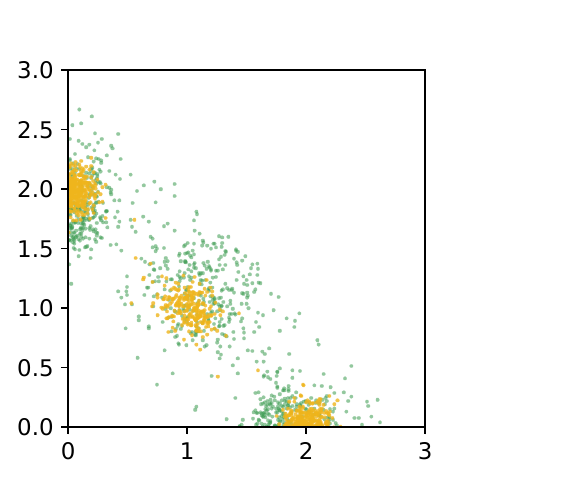}
\caption{Multi-stable: samples}
\end{subfigure}
\par\smallskip
\includegraphics[width=0.65\textwidth]{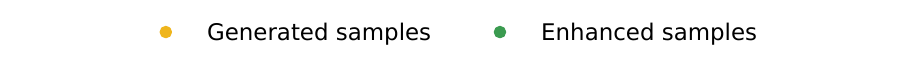}
\caption{Learned potentials and self-generated samples for the two-dimensional
benchmarks. Columns: double well ($D=0.05$), limit cycle ($D=0.1$), and the multi-stable model ($D=0.01$). (a--c) Filled contours of the learned potential
$U_\theta=-D\log q_\theta$, shifted to a minimum of zero.
Dark dots are direct flow samples. White and gray arrows show
$-\nabla U_\theta$ and $F+\nabla U_\theta$, respectively, using a common scale
within each system to preserve relative magnitudes. Arrows are shown in
low-potential regions. (d--f) Generated samples (yellow) and enhanced HJB collocation samples
(green). Both sets are generated from the final model specifically for visualization.
Enhancement uses biased dynamics with energy-window filtering for the double
well and multi-stable systems, and Gaussian perturbations for the limit cycle, following
the corresponding training configurations.}
\label{fig:landscapes-lowd}
\end{figure}

The third system is a two-variable, three-attractor model of cell-state
regulation \citep{Wang11,zhao2024nsr}, at $D=0.01$. Its
separated basins test mode discovery, coverage of the low-density
regions, and the ability to recover transition regions at small noise.  The reference is
obtained from the same stationary Fokker--Planck discretization used for
the EPR-Net comparison. Enhanced EPR achieves an rRMSE of
$0.0189$, compared with $0.6876$ for the HJB-only control
(\cref{tab:low-d-results}; \cref{fig:landscapes-lowd}(c,f)).
The Enhanced EPR landscape separates three low-potential basins along the diagonal region, with generated samples occupying these basins. The HJB-only comparison shows a broader low-potential region, while KL captures the main low-potential regions with less regular contours (\cref{fig:lowd-method-comparison}). The results illustrate how the learned density brings basin identification and direct sampling together in a small-noise multistable system. Run-to-run variability is reported in \cref{tab:low-d-results}.

\subsection{Twelve-dimensional multistable Gaussian-mixture model}
\label{sec:exp-gmm}

The first high-dimensional example is a two-component Gaussian-mixture model on
$\R^{12}$ constructed so that its normalized stationary density, and hence
its exact finite-noise landscape, are known \citep{zhao2024nsr}.
An antisymmetric component in the drift introduces stationary circulation
while preserving the prescribed mixture density. This controlled problem
tests whether EPR-NF can preserve both modal masses and the barrier between
them while learning a full-dimensional normalized density on an unbounded
support. We evaluate the bulk landscape using $65\,536$ samples from the
analytic mixture. The reference potential is shifted by its minimum over the
sample set, and points below $20D$ are retained. The learned potential is
then aligned using the median prediction--reference difference on these
points. The resulting rRMSE measures potential error weighted by the reference
distribution in this energy range. At $D=0.1$, the bulk rRMSE is $0.0117$,
the learned-to-reference barrier
ratio is $0.997$, and the two-mode weight $L_1$ error is $0.020$.
\Cref{fig:12d-line} shows the
landscape along the line joining the two component means.
These results demonstrate accurate reconstruction of the bulk landscape,
modal weights, and interwell barrier in this controlled high-dimensional
benchmark. The two components share an isotropic covariance, so
multimodality lies along the line joining their means, with Gaussian
structure in the orthogonal directions. Together, these measures demonstrate that the full-dimensional density provides a consistent basis for landscape reconstruction and direct sampling.

\subsection{Full-dimensional biological networks and reduced landscapes}

For each biological network, we first train a full-dimensional EPR-NF model
from the known drift and diffusion coefficient. Following
\cref{sec:model-reduction}, we freeze this model, draw samples to fit the
conditional drift in the retained coordinates, and then train a reduced
flow with the fitted drift. The reduced flow provides both a landscape
and a direct sampler. All training stages use model-generated samples;
simulated stationary samples projected onto the retained coordinates provide the empirical marginal landscapes used for evaluation. Structural correlation measures agreement in landscape variation across the evaluation region, while aligned rRMSE measures potential differences after fitting an additive offset. \Cref{fig:dr-landscapes} shows the resulting landscapes
and model-generated samples.

For evaluation, projected stationary samples are binned into histograms,
smoothed with a Gaussian kernel of one grid-cell standard deviation, and
normalized. Densities are floored at $e^{-50}$ times their maximum before
applying $-D\log$; each resulting reference landscape is minimum-shifted. A constant offset is fitted
by the median of model minus reference potential on reference cells below
$20D$; correlation and aligned rRMSE are then evaluated below $30D$.
The DR evaluation uses these grids, masks, and additive offsets.
The references are used to assess configurations and report reconstruction
metrics. Training losses use model-generated samples.

\subsubsection{Three-dimensional cell-cycle model}
\label{sec:exp-dr3}

\begin{figure}[t]
\centering
\begin{subfigure}[t]{0.48\textwidth}
\centering
\includegraphics[width=\linewidth]{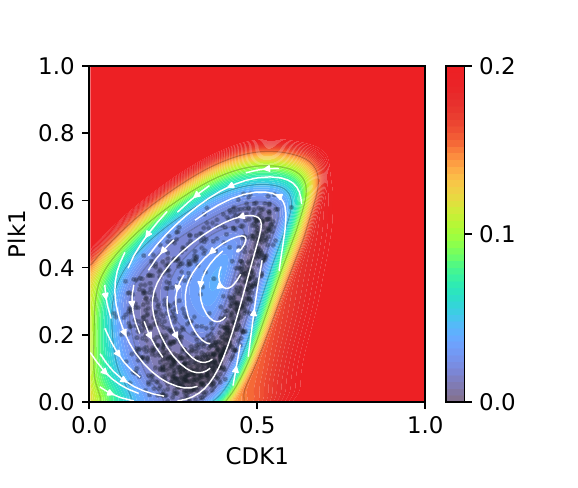}
\caption{3D cell-cycle model}
\end{subfigure}
\hfill
\begin{subfigure}[t]{0.48\textwidth}
\centering
\includegraphics[width=\linewidth]{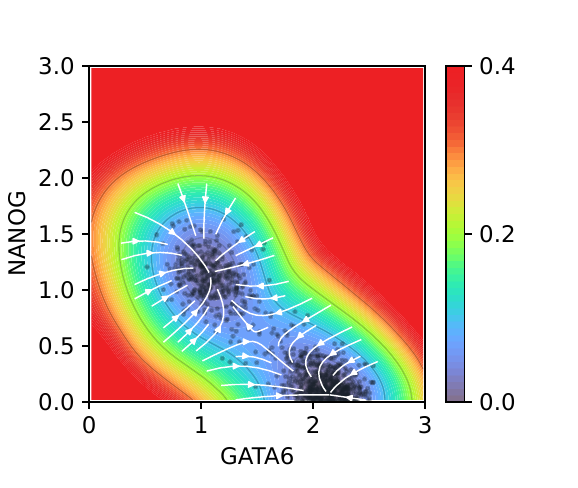}
\caption{52D gene-regulatory network}
\end{subfigure}
\caption{Reduced landscapes of biological networks: (a) the 3D cell-cycle system at $D=0.01$; (b) the 52D gene network at $D=0.02$. Dark dots are projected samples from the corresponding full-dimensional flow. White streamlines show the learned conditional drift $g_\eta$ in low-potential regions. Colors represent $V-\min V$. Training configurations are described in \ref{app:dimension-reduction}.}
\label{fig:dr-landscapes}
\end{figure}

The three-variable model of Ferrell et al.~\citep{Ferrell11} is used to assess
reconstruction of an oscillatory stationary marginal in CDK1--Plk1 coordinates. At $D=0.01$,
a flow is trained on all three variables. The empirical reference uses
$200\,000$ projected stationary samples on a $161\times161$ grid in $[0,1]^2$. The full-flow samples projected onto these coordinates trace the
oscillatory region, and the DR landscape recovers an elongated valley. Projected samples follow this low-potential region, while the learned conditional-drift streamlines trace the associated circulation (\cref{fig:dr-landscapes}(a)).
DR reaches a structural correlation of $0.9352$ and an aligned rRMSE of
$0.2942$, averaged over three reduced fits. Agreement is strongest in the region containing most of the probability mass;
the boundary of the empirical reference differs from the smooth learned contours. The result shows that model-generated samples can support a reduced landscape that captures the dominant oscillatory organization.

\subsubsection{Fifty-two-dimensional gene-regulatory system}
\label{sec:exp-gene-52d}

For the gene-regulatory network \citep{Li13,zhao2024nsr}, the first flow
represents all $52$ variables, and reduction retains GATA6 and NANOG.
The empirical reference uses $100\,000$ projected stationary samples on an
$81\times81$ grid in $[0,3]^2$. At $D=0.02$, samples from the full-dimensional flow are used with the projected drift to fit the conditional force before training the reduced flow. Their projections occupy both principal phenotypic basins in the GATA6--NANOG plane, and the DR landscape preserves their separation (\cref{fig:dr-landscapes}(b)). DR attains a
mean correlation of $0.9143$ and a mean aligned rRMSE of $0.2850$
over three reduced fits, while providing
a normalized two-dimensional density and a direct sampler. This example demonstrates that the model-based workflow can reveal interpretable low-dimensional state organization in a high-dimensional network.

These experiments establish a workflow in which full-dimensional training,
force estimation, and reduced landscape learning all use model-generated
samples. The resulting two-dimensional landscapes recover the oscillatory
valley of the cell-cycle model and the separated phenotypic basins of the
gene network.

\FloatBarrier

\section{Discussion}
\label{sec:discussion}

Generative models provide a flexible framework for distribution learning and
sampling \citep{liu2025scoresingularity,liu2026rddpm,liu2026fesfm}.
EPR-NF changes the role of sampling in landscape construction.  Samples no
longer constitute observations of the target stationary density; instead,
they are collocation points proposed by a normalized PDE solver.  The same
flow parameters provide a density, a score, a landscape, and a sampler.
Score matching estimates statistical models through log-density gradients
\citep{hyvarinen05a}, and score-based generative models use learned gradients
to drive sampling \citep{Song2019}. In EPR-NF, the score is computed from a
normalized density trained against the prescribed drift and stationary PDE.

Adding EPR to HJB substantially improves landscape reconstruction for the
limit-cycle and multi-stable systems under matched protocols; HJB alone
already achieves high accuracy on the double well. Enhanced EPR
also produces lower landscape errors than the sample-based KL baseline
on all three two-dimensional systems. The landscape metrics assess the
reconstruction of basins and transition regions, including regions with
little stationary mass. The double-well ablations in
\ref{app:ablations} further show that enhanced collocation and diffusion
continuation materially improve accuracy under the tested settings.

The two terms play distinct roles. EPR projects the drift under the
current frozen sampling measure, and its self-bootstrap update admits the
local convergence analysis in \cref{sec:theory}. HJB supplies an additional
pointwise stationary constraint whose collocation can be enriched in
low-density regions. Enhanced EPR uses both mechanisms; the pure
EPR theorem isolates the mathematical basis of its self-sampling component.

The analysis connects local stability of self-bootstrap EPR to the
stationary current and the weighted Neumann inverse. It gives a sufficient
small-current condition for geometric convergence and quantifies the
accuracy of inexact updates, linking the force-projection formulation
to a convergent population iteration.

The full-dimensional and reduced models also provide a basis for further
development. Adaptive sampling can be investigated to improve coverage
of high-dimensional landscapes, while efficient second-derivative
calculations can reduce the cost of the stationary residual. Extending
the formulation to general diffusion tensors would broaden its
application to anisotropic stochastic systems.

\section{Conclusion}
\label{sec:conclusion}

We presented EPR-NF, a normalized formulation with model-based sampling for learning
finite-noise non-equilibrium landscapes from a prescribed drift. Its Enhanced EPR
objective couples EPR force projection with a pointwise stationary HJB constraint,
using the same flow to represent the density and generate training points.
For pure self-bootstrap EPR, we identify the unique stationary fixed
point and prove local geometric convergence under an explicit stability
condition, together with an inexact-update error bound. The numerical study demonstrates accurate reconstruction on several
multistable benchmarks and extends the workflow to full-dimensional
biological networks. Model-generated samples support reduced-force
learning and the subsequent landscape reconstruction, so stationary trajectory data are no
longer required at any training stage of this workflow. The reduced
landscapes recover the principal oscillatory and phenotypic structures.

\clearpage

\appendix

\section{Theoretical Derivations and Proofs}
\label{app:theory}

\subsection{Proofs for Pure Self-Bootstrap EPR}
\label{app:proofs}

\paragraph{Frozen-measure minimization and the fixed point}
For a positive continuous $r$ on $\overline\Omega$, minimization of
$\mathcal L(V;r)$ over $H^1$ modulo constants is a coercive quadratic
problem. Weighted Poincar\'e and Lax--Milgram give a unique solution after
imposing $\int pV=0$. Its first variation is
\begin{equation}
 D_V\mathcal L(V;r)[\phi]
   =2\int_\Omega r(F+\nabla V)\cdot\nabla\phi\,\mathrm{d}x,
 \qquad \phi\in H^1(\Omega).
 \label{eq:frozen-epr-variation}
\end{equation}
The density is not differentiated. Integration by parts gives
\eqref{eq:weighted-el}, including its natural no-flux boundary condition.
Conversely, a weak solution of that equation is the unique minimizer in
the gauge, since the difference in objectives is
$\int r|\nabla(V-V_*)|^2$.

\begin{proof}[Proof of \cref{prop:fixed-point}]
Put $v=F+\nabla U$. Stationarity and reflection give
$\nabla\cdot(pv)=0$ and $pv\cdot n=0$. Hence
$\int pv\cdot\nabla\phi=0$ for every $\phi\in H^1$.
Expanding $F+\nabla V=v+\nabla(V-\bar U)$ eliminates the cross term and
gives \eqref{eq:epr-projection-bootstrap}. Equality holds only when
$V-\bar U$ is constant; the gauge sets that constant to zero.
\end{proof}

\begin{proof}[Proof of \cref{thm:selfconsistent}]
Set $r=q_V$ in \eqref{eq:frozen-epr-variation} after taking the derivative.
Since $q_V\nabla V=-D\nabla q_V$, stationarity reads
\[
 \int_\Omega(q_VF-D\nabla q_V)\cdot\nabla\phi\,\mathrm{d}x=0
 \qquad\text{for every }\phi\in H^1(\Omega).
\]
This is the weak reflecting stationary Fokker--Planck equation. The assumed
uniqueness of its normalized solution gives $q_V=p$. A fixed point of an
exact EPR update satisfies this identity, and the preceding proposition
supplies the converse. For an exact iteration converging in the stated potential norm, passing
to the limit in its weak variational equation gives the same
self-consistency condition.
\end{proof}

\paragraph{The weighted Neumann inverse}

Let $\mathcal Y=\{f\in C^{0,\alpha}:\int pf=0\}$. The operator
$\mathcal A:\mathcal X\to\mathcal Y$ in \eqref{eq:epr-neumann} is a
bounded bijection with bounded inverse. To see this, solve
\[
 \int_\Omega p\nabla u\cdot\nabla\phi\,\mathrm{d}x
       =\int_\Omega pf\phi\,\mathrm{d}x,\qquad \int pu=0.
\]
The positivity of $p$ and weighted Poincar\'e inequality yield a unique
weak solution. Its Neumann regularity gives $u\in C^{2,\alpha}$.
For completeness, writing
$\mathcal A=-\Delta-\nabla\log p\cdot\nabla$, the Neumann Schauder estimate and
interpolation give
\[
 \|u\|_{C^{2,\alpha}}
   \le C\bigl(\|\mathcal Au\|_{C^{0,\alpha}}+\|u\|_{C^0}\bigr).
\]
This is the usual lower-order perturbation of the Neumann Laplacian
estimate \citep{nardi2014neumann}. The $C^0$ term can be removed on
$\mathcal X$: otherwise normalize a sequence to unit $C^{2,\alpha}$ norm
with $\mathcal Au_j\to0$. Compactness in $C^2$ and the estimate give a
nonzero limiting kernel element. But integration against $pu$ gives
$\int p|\nabla u|^2=0$; connectedness and the gauge force $u=0$, a
contradiction. This proves the inverse bound. The same argument applies to
smooth positive nearby weights. In particular
$\mathcal R=\mathcal A^{-1}P_p$ is bounded from $C^{0,\alpha}$ to
$\mathcal X$.

\paragraph{An exact local error equation}

Write $W=\bar U+h$ and seek the updated potential as $V=\bar U+g$,
where $h,g\in\mathcal X$. Direct substitution into the weighted equation
gives
\begin{equation}
 q_W^{-1}\nabla\cdot\{q_W(F+\nabla V)\}
 =-\mathcal Ag-\frac1D(v+\nabla g)\cdot\nabla h.
 \label{eq:epr-strong-error}
\end{equation}
Here the zeroth-order term cancels because $\nabla\cdot(pv)=0$.
Moreover $\int p\,v\cdot\nabla h=0$, by integration by parts.
Applying $\mathcal A^{-1}P_p$ to the zero-residual equation gives
\[
 (I+\mathcal K_h)g=\mathcal B h,
 \qquad \|\mathcal K_h\|\le c\|h\|_{\mathcal X}.
\]
If $c\|h\|_{\mathcal X}<1$, the inverse exists by the Neumann series,
and defines a unique $g\in\mathcal X$.

The projection has not lost an equation. For this constructed $g$,
\eqref{eq:epr-strong-error} is a spatial constant $a$, since its $P_p$
projection vanishes. Multiply by $q_W$ and integrate. The left-hand side
integrates to the boundary flux, which is zero because
$(v+\nabla g)\cdot n=0$. Since $\int q_W=1$, this implies $a=0$.
Thus $V$ solves the full weighted Neumann equation and is the unique
EPR minimizer. We have proved \eqref{eq:epr-error-identity}, and also
\begin{equation}
 \|\mathcal S(\bar U+h)-\bar U\|_{\mathcal X}
 \le\frac{b}{1-c\|h\|_{\mathcal X}}\|h\|_{\mathcal X}.
 \label{eq:epr-radial-bound}
\end{equation}
The same identity yields
$\mathcal S(\bar U+h)-\bar U=\mathcal B h+O(\|h\|_{\mathcal X}^2)$,
so \eqref{eq:epr-linearization} is the Fr\'echet derivative, not merely
a formal linearization. The formula also makes the update continuous in
this neighborhood.

The current dependence can also be read from the energy identity
\[
 \int p\nabla(\mathcal B h)\cdot\nabla\phi
       =\frac1D\int ph\,v\cdot\nabla\phi.
\]
If $C_P$ is the weighted Poincar\'e constant, this implies
$\|\nabla\mathcal B h\|_{L^2(p)}
 \le D^{-1}\sqrt{C_P}\|v\|_\infty\|\nabla h\|_{L^2(p)}$.
The convergence theorem uses the stronger H\"older norm bound $b<1$;
the energy estimate describes the role of circulation but is not
substituted for that hypothesis.

\paragraph{Local and inexact convergence}

\begin{proof}[Proof of \cref{thm:bootstrap-convergence}]
Choose $r$ as in the theorem, which is possible whenever $b<1$.
For $\|h\|_{\mathcal X}\le r$, \eqref{eq:epr-radial-bound} is at most
$\kappa\|h\|_{\mathcal X}$. Hence the ball is invariant. Induction gives
\eqref{eq:epr-local-rate}. Exponentiation, integration, and normalization
are smooth on this bounded H\"older neighborhood, so the densities and
their normalized potentials converge in $C^{2,\alpha}$ as claimed.
The sufficient condition \eqref{eq:epr-small-current} follows immediately
by applying the inverse and product bounds to \eqref{eq:epr-linearization}.
If $v=0$, the integrand at $U$ is zero for every positive weight $r$;
strict convexity in the gradient identifies its minimizer as $U+C$.
\end{proof}

\begin{proof}[Proof of \cref{cor:epr-inexact}]
The radial estimate gives
$\|W_{k+1}-\bar U\|\le\kappa\|W_k-\bar U\|+\eta_k$.
The stated upper bound on $\eta_k$ preserves the radius $r$.
Iterating this scalar inequality proves \eqref{eq:epr-inexact-bound}.
If $\eta_k\to0$, split the convolution into a finite prefix, which decays
geometrically, and a tail bounded by
$\sup_{j\ge J}\eta_j/(1-\kappa)$. First let $k\to\infty$, then
$J\to\infty$. A uniform error floor instead gives the stated limit superior.
\end{proof}

\section{Supplementary Experiments and Implementation}
\label{app:experiments}

\begin{figure}[h]
\centering
\includegraphics[width=0.42\textwidth]{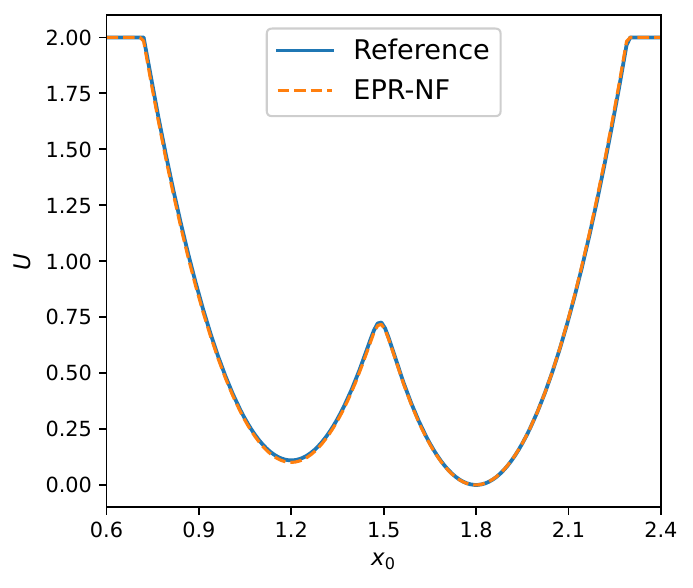}
\caption{Twelve-dimensional Gaussian mixture: landscape along the line
joining the two component means: the reference and final
EPR-NF landscapes. This one-dimensional section illustrates the well
and barrier geometry; modal masses are assessed separately from direct
samples.}
\label{fig:12d-line}
\end{figure}

This section collects all supplementary experimental and implementation
material: training diagnostics, the Bernstein flow implementation, the
weak-form comparison, training protocols, reduced-force regression, and
ablation studies. Theoretical derivations and proofs are grouped in
\ref{app:theory}.
\Cref{fig:12d-line} is the twelve-dimensional landscape along the line
joining the two component means.

\subsection{Bernstein autoregressive flow}
\label{app:bnf}

The experiments use a Bernstein autoregressive flow
\citep{ramasinghe2022bernstein,durr2024bernstein,amorese2026bernstein}.
On the unit cube, each of $K$ blocks transforms coordinate $k$ by a scalar
monotone map
\begin{equation}
  y_k=(1-\eta)\,B\!\left(u_k;\,w(u_{<k})\right)+\eta\,u_k,
  \qquad
  B(u;w)=\sum_{i=0}^{M} c_i\binom{M}{i}u^i(1-u)^{M-i},
  \label{eq:bernstein-map}
\end{equation}
where the coefficients $c_i$ are the cumulative sums of a softmax, so
$0=c_0\le\cdots\le c_M=1$ and $B$ is a Bernstein-polynomial CDF.  The first
coordinate has independently parameterized logits; logits for subsequent
coordinates are generated by a network conditioned on the preceding coordinates.  Coordinate order is reversed between blocks.  The
log-density is the sum of the log diagonal Jacobian entries.  Inversion has
no convenient closed form and is performed sequentially by coordinatewise
bisection.  Analytic first derivatives of this map, used for the log-determinant,
are part of the same family
\citep{ramasinghe2022bernstein,amorese2026bernstein}.  The identity weight
$\eta>0$ only mixes $B$ with the identity and does not enlarge the family.
For bounded problems, an affine map transforms the physical box to the unit
cube, with a uniform base distribution ensuring support on the physical box
\citep{amorese2026bernstein}.  For the twelve-dimensional mixture, the same
core transformation is combined with a standard-normal CDF.  Degree, depth, conditioner width, $\eta$,
and the bisection tolerance are numerical choices
(\ref{app:experimental-details}).

We make two implementation modifications to train this flow with the EPR
and HJB losses, while retaining the Bernstein map.

\paragraph{Forward jet of the log-density}
Here, a jet denotes a mathematical representation of a value and its
spatial derivatives, rather than a network layer or a software-specific
construction. If an intermediate scalar \(v\) depends on the spatial input \(x\), its
second-order jet at \(x\) is represented by the triple
\begin{equation}
  \bigl(v(x),\;\nabla_x v(x),\;\Delta_x v(x)\bigr).
  \label{eq:jet}
\end{equation}
The first entry is the value.  The second is the gradient of that value
with respect to the physical coordinate \(x\).  The third is the Laplacian,
the trace of the second-derivative matrix in the same coordinate.  These
are derivatives in \(x\).  They are distinct from derivatives with respect
to the parameters \(\theta\).

The losses require these quantities for the scalar log-density.  The EPR term uses the score \(\nabla\log q_\theta\), and the
HJB residual also uses the Laplacian \(\Delta\log q_\theta\).  Because
\(\log q_\theta\) contains a log-determinant, both quantities depend on
derivatives of the scalar Bernstein map beyond the first derivative already
used to evaluate the density.

One way to obtain them is to first evaluate the scalar
\(\log q_\theta(x)\) and then apply the chain rule to that scalar, once for
the gradient and again for the Laplacian.  That procedure, often called
automatic differentiation, is a valid evaluation of the same derivatives.
During training, we instead propagate these derivatives alongside the forward evaluation. As the network is evaluated from the
input \(x\) toward \(\log q_\theta\), every intermediate scalar is stored
as a jet \eqref{eq:jet} and updated at each operation using the product rule and
the chain rule,
\begin{align}
  \Delta(fg)
  &=f\,\Delta g+g\,\Delta f+2\nabla f\cdot\nabla g,\\
  \Delta h(f)
  &=h'(f)\,\Delta f+h''(f)\,\|\nabla f\|^2.
\end{align}
The same two identities are applied, in closed form, through each linear
layer, through the exact GELU activation of the conditioning network, through the softmax, the
Bernstein basis, and the logarithm, and through the composition of blocks.
The conditioning network has two hidden layers with the GELU activation so that the
first and second derivatives of every activation exist in closed form.
Finite differences and quadrature are not used to approximate the monotone
map.  The resulting derivatives are the analytic derivatives of this finite
implementation.  These evaluations remain subject to floating-point rounding and the
clipping described below.

After the jet has produced \(\log q_\theta\), its score, and its Laplacian,
the training loss is an ordinary function of those three quantities and of
\(\theta\).  The derivative of the loss with respect to \(\theta\) is then
the usual chain-rule derivative of that function.  Applying the chain rule
directly to the evaluated scalar \(\log q_\theta\) is retained as a numerical
check: it must reproduce the jet's value, score, Laplacian, and parameter
derivative.  The reported training runs use the jet.

Sampling does not pass through this jet.  Base samples are drawn, the
blocks are inverted by bisection, and the closure of each sample is checked
by a forward evaluation.  Sampling is performed without gradient tracking. Gradients
are obtained by evaluating the density and its jet at the sampled points.

\paragraph{Finite-precision agreement of the density and its derivatives}

In exact arithmetic the cumulative softmax coefficients satisfy
$B(0)=0$ and $B(1)=1$, and a positive derivative makes the scalar inverse
unique.  In floating-point arithmetic, the right endpoint may deviate from $1$,
and rounding may produce CDF values slightly outside $[0,1]$.

The cumulative coefficients are therefore normalized again so that the
right endpoint is exactly one, and CDF values outside $[0,1]$ due to rounding
are clipped to that interval.  The map used for $\log q_\theta$ and
the map differentiated by the jet apply this same clipping, including the
derivative of the clip.  The Jacobian used to evaluate the density therefore corresponds to
the function evaluated in the implementation.  Without that
agreement, the HJB residual would be the residual of a different map from
the one whose normalized density supplies the landscape and the EPR samples.

\subsection{Training dynamics}
\label{app:gmm-check}
\begin{figure}[t]
\centering
\includegraphics[width=\textwidth]{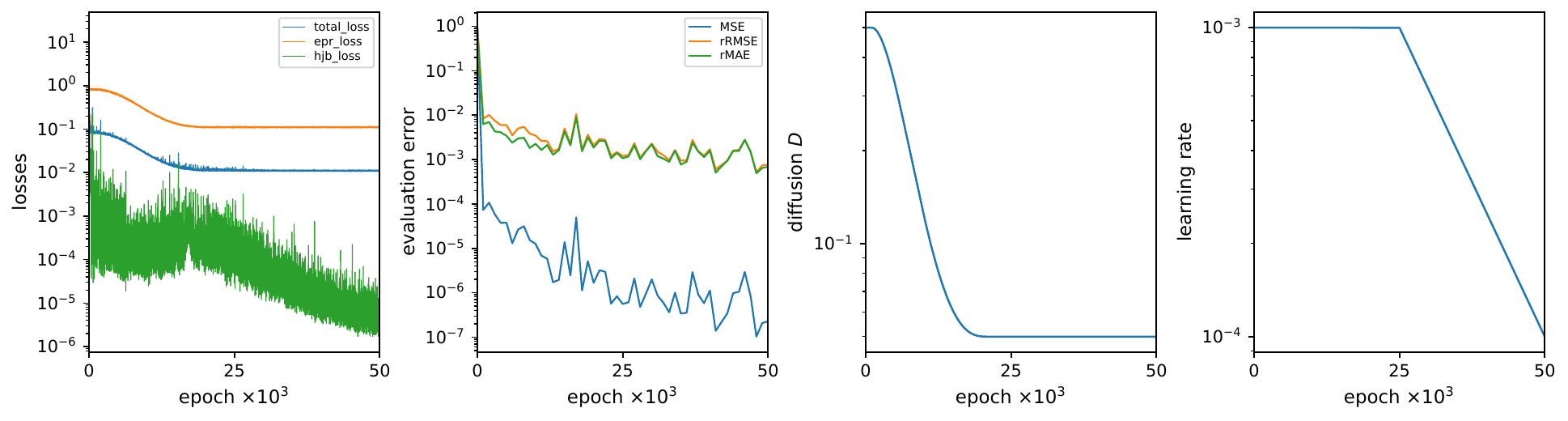}
\caption{Training dynamics of the final double-well run: the EPR
and HJB losses, the monitored evaluation errors, the anchored-geometric
diffusion schedule $D_k$ from $10D$ to $D$, and the learning rate.}
\label{fig:training}
\end{figure}

\Cref{fig:training} shows the evolution of the double-well training
losses and evaluation errors alongside the diffusion and learning-rate
schedules.


\subsection{Landscape comparison across training objectives}
\label{app:lowd-method-comparison}

\begin{figure}[p]
\centering
\begin{subfigure}[t]{0.32\textwidth}
\centering
\includegraphics[width=\linewidth]{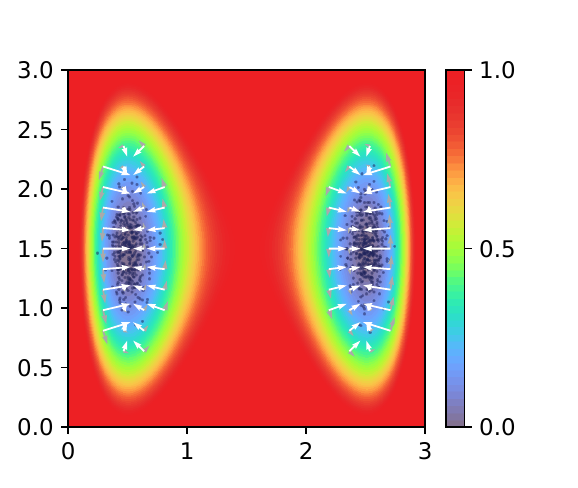}
\caption{Double well: Enhanced EPR}
\end{subfigure}
\hfill
\begin{subfigure}[t]{0.32\textwidth}
\centering
\includegraphics[width=\linewidth]{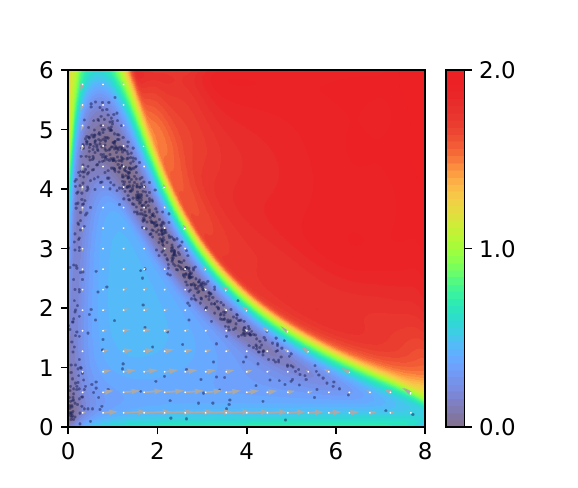}
\caption{Limit cycle: Enhanced EPR}
\end{subfigure}
\hfill
\begin{subfigure}[t]{0.32\textwidth}
\centering
\includegraphics[width=\linewidth]{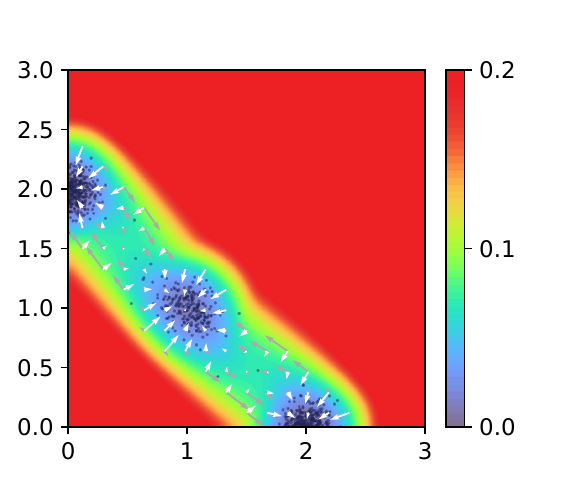}
\caption{Multi-stable: Enhanced EPR}
\end{subfigure}
\par\smallskip
\begin{subfigure}[t]{0.32\textwidth}
\centering
\includegraphics[width=\linewidth]{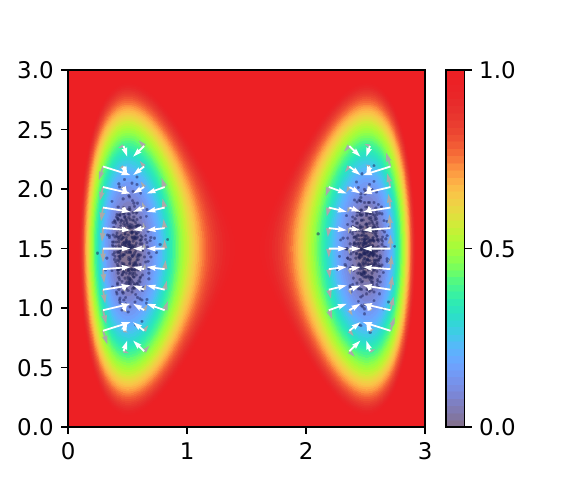}
\caption{Double well: HJB only}
\end{subfigure}
\hfill
\begin{subfigure}[t]{0.32\textwidth}
\centering
\includegraphics[width=\linewidth]{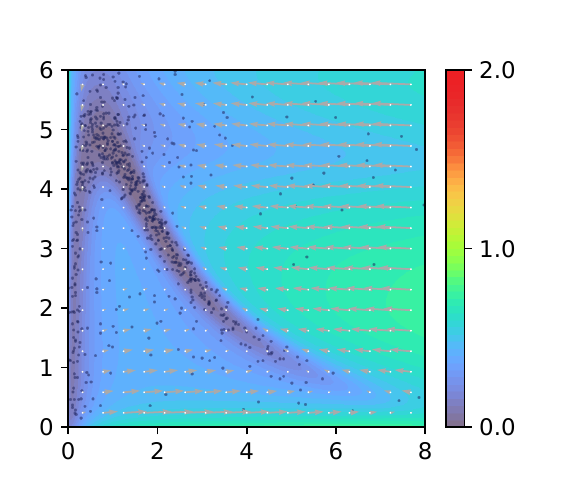}
\caption{Limit cycle: HJB only}
\end{subfigure}
\hfill
\begin{subfigure}[t]{0.32\textwidth}
\centering
\includegraphics[width=\linewidth]{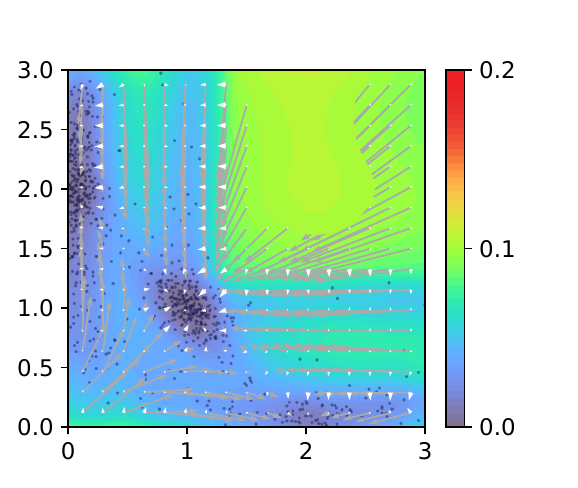}
\caption{Multi-stable: HJB only}
\end{subfigure}
\par\smallskip
\begin{subfigure}[t]{0.32\textwidth}
\centering
\includegraphics[width=\linewidth]{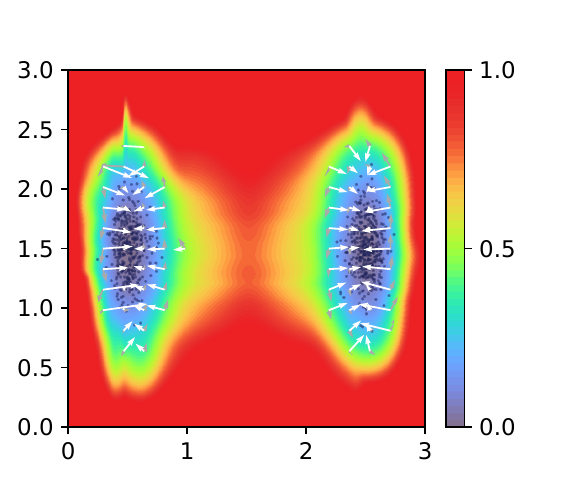}
\caption{Double well: KL}
\end{subfigure}
\hfill
\begin{subfigure}[t]{0.32\textwidth}
\centering
\includegraphics[width=\linewidth]{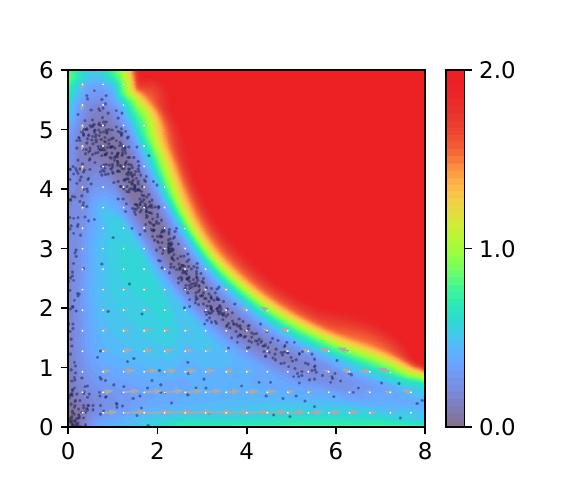}
\caption{Limit cycle: KL}
\end{subfigure}
\hfill
\begin{subfigure}[t]{0.32\textwidth}
\centering
\includegraphics[width=\linewidth]{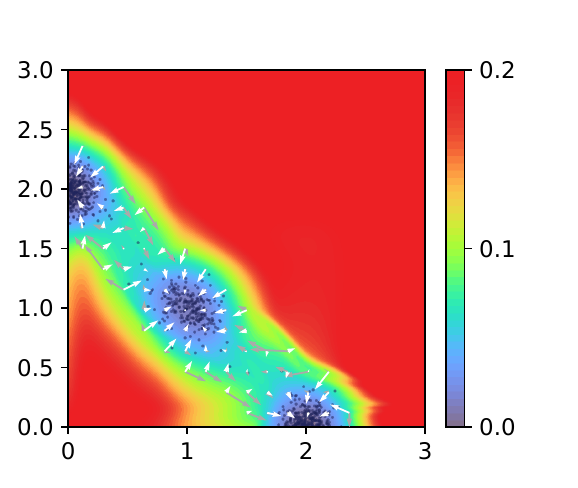}
\caption{Multi-stable: KL}
\end{subfigure}
\caption{Final learned potentials for the two-dimensional benchmarks. Columns show
the double-well ($D=0.05$), limit-cycle ($D=0.1$), and multi-stable ($D=0.01$)
systems; rows show Enhanced EPR, HJB only, and KL.
All panels display $U_\theta-\min U_\theta$, with a separate colorbar for
each panel and matching color scales and coordinate ranges within each system.
Dark dots are direct samples from the model shown in each panel. White
arrows represent $-\nabla U_\theta=D\nabla\log q_\theta$; gray arrows
represent $F+\nabla U_\theta$. Both components share one arrow scale across
the three methods for each system, using the scale of \cref{fig:landscapes-lowd},
and are displayed in low-potential regions. The Enhanced EPR potentials and samples are the same as those in
\cref{fig:landscapes-lowd}.}
\label{fig:lowd-method-comparison}
\end{figure}

\Cref{fig:lowd-method-comparison} compares the final potentials obtained by
Enhanced EPR, HJB-only training, and KL for the three two-dimensional
benchmarks in \cref{tab:low-d-results}. Enhanced EPR and HJB-only training
give closely matching double-well landscapes. The limit-cycle and
multi-stable panels reveal differences in the valley and basin geometry,
while KL captures the main low-potential regions with less regular
contours.

\FloatBarrier
\subsection{Comparison with a weak-form simulation-free sampler}
\label{sec:exp-wgs}

The weak generative sampler (WGS) of Cai et al.~\citep{cai2026wgs} provides
a related simulation-free approach. It trains a generator
$x=G(z)$ so that its pushforward satisfies the weak form of the stationary
Fokker--Planck equation against a family of Gaussian radial-basis test
functions centered at jittered samples.  Like EPR-NF it requires no
simulated trajectories.  Unlike EPR-NF, it constrains a weak integral
identity rather than a pointwise equation. The weak formulation does not
require a tractable model density; when implemented with an invertible flow,
however, density evaluation and landscape reconstruction remain available.

We compare these methods on the double well at $D=0.1$ using the same diffusion
schedule as the EPR-only run, counting one WGS update as one epoch.  Two
WGS variants are compared: the official float64 RealNVP
generator with its Gaussian radial-basis tests and soft boundary penalty,
and the same objective with the Bernstein flow of
\ref{app:bnf} as the generator and Neumann heat-kernel
tests adapted to the reflecting box.  Sample counts, kernel width, and
optimizer settings are recorded in \ref{app:experimental-details}.
All methods are evaluated using rRMSE and rMAE on the same masked reference
grid, following the definitions in \eqref{eq:rmse} and~\eqref{eq:metrics}.

\Cref{tab:wgs} reports larger landscape errors for both tested WGS
variants than for EPR-NF on this benchmark. The Bernstein variant has lower
rRMSE and rMAE than the RealNVP variant. Across five additional RealNVP configurations, each evaluated in one run, the mean rRMSE is $2.1713$.

\begin{table}[t]
\centering
\caption{Comparison with WGS on the double well at $D=0.1$ under the
identical annealing schedule. Results at the final training epoch are reported
as means with standard deviations.
EPR-NF here uses the EPR
term alone, with $\lambda=0$.}
\label{tab:wgs}
\begin{tabular}{lcc}
\toprule
Method & rRMSE & rMAE \\
\midrule
WGS, official RealNVP & \withsd{0.8173}{0.1007} & \withsd{0.5937}{0.0624} \\
WGS, Bernstein flow & \withsd{0.4223}{0.0504} & \withsd{0.3538}{0.0368} \\
EPR-NF, EPR only & \withsd{0.0060}{0.0020} & \withsd{0.0049}{0.0014} \\
\bottomrule
\end{tabular}
\end{table}

\begin{figure}[H]
\centering
\begin{subfigure}[t]{0.32\textwidth}
\centering
\vspace{0pt}
\includegraphics[width=\linewidth]{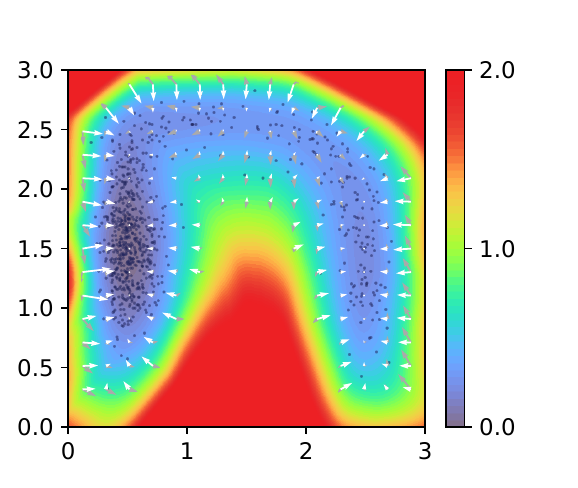}
\caption{WGS: RealNVP}
\end{subfigure}
\hfill
\begin{subfigure}[t]{0.32\textwidth}
\centering
\vspace{0pt}
\includegraphics[width=\linewidth]{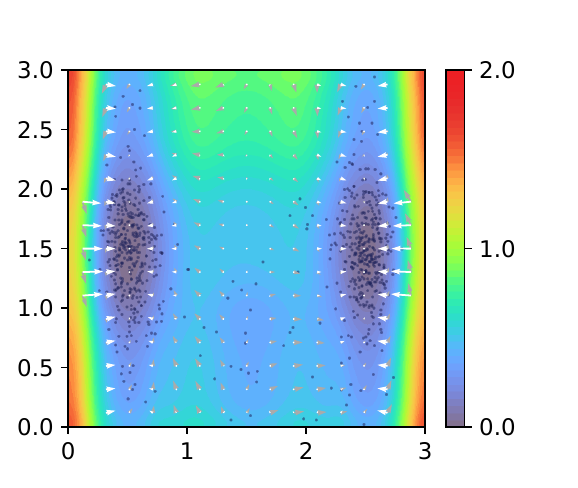}
\caption{WGS: Bernstein}
\end{subfigure}
\hfill
\begin{subfigure}[t]{0.32\textwidth}
\centering
\vspace{0pt}
\includegraphics[width=\linewidth]{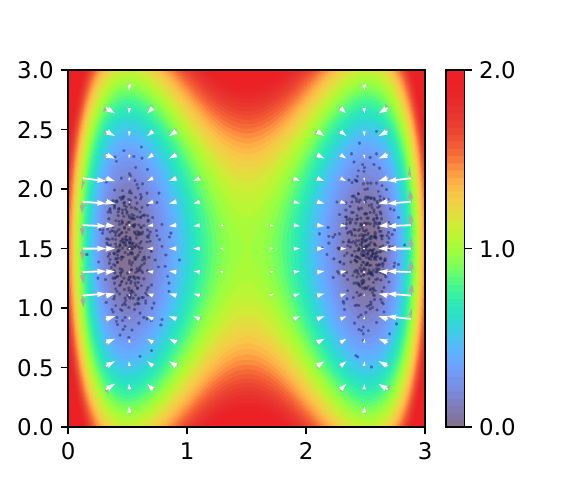}
\caption{EPR-NF: EPR only}
\end{subfigure}
\caption{Endpoint models for the three configurations in \cref{tab:wgs}
on the double well at $D=0.1$: (a) WGS with the official RealNVP generator,
(b) WGS with the Bernstein flow, and (c) EPR-NF trained with EPR alone.
Filled contours show each learned potential shifted to a minimum of zero,
with the same color scale and a separate colorbar in each panel. Dark dots are samples
generated directly by the corresponding model. White arrows show
$-\nabla U_\theta=D\nabla\log q_\theta$; gray arrows show
$F+\nabla U_\theta$. All panels and both force components use the same
arrow scale, with arrows shown in low-potential regions. The panels show individual final models, whereas
the table reports aggregate metrics.}
\label{fig:wgs-landscapes}
\end{figure}

\Cref{fig:wgs-landscapes} complements \cref{tab:wgs} with the learned
landscape, direct samples, and force decomposition of each method.
The landscape errors assess reconstruction of the geometry within and
between the wells.

\subsection{Training protocol}
\label{app:experimental-details}

\begin{figure}[t]
\centering
\begin{subfigure}[t]{0.32\textwidth}
\centering
\includegraphics[width=\linewidth]{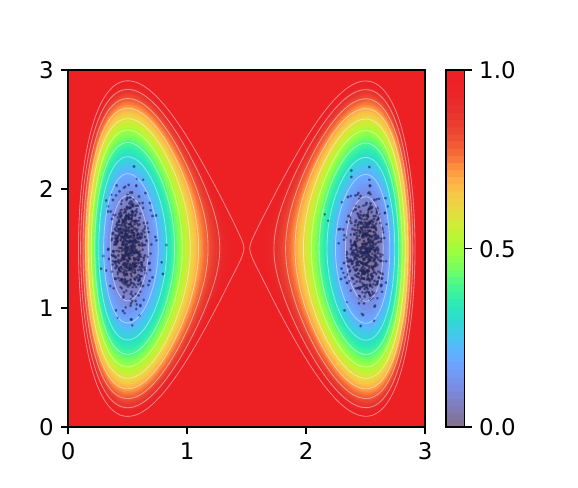}
\caption{Flow samples, $\sigma=0$.}
\end{subfigure}
\hfill
\begin{subfigure}[t]{0.32\textwidth}
\centering
\includegraphics[width=\linewidth]{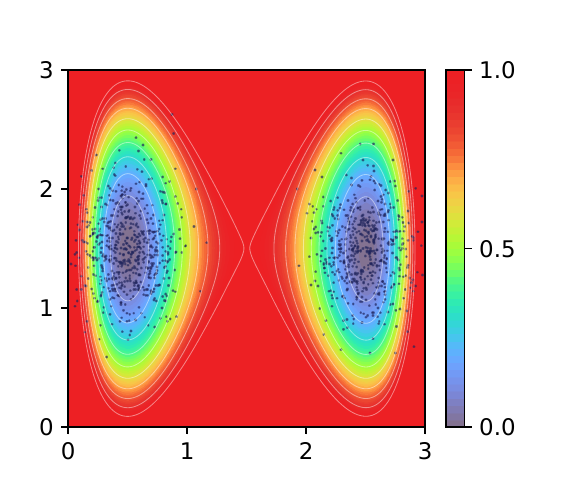}
\caption{Gaussian noise, $\sigma=0.2$.}
\end{subfigure}
\hfill
\begin{subfigure}[t]{0.32\textwidth}
\centering
\includegraphics[width=\linewidth]{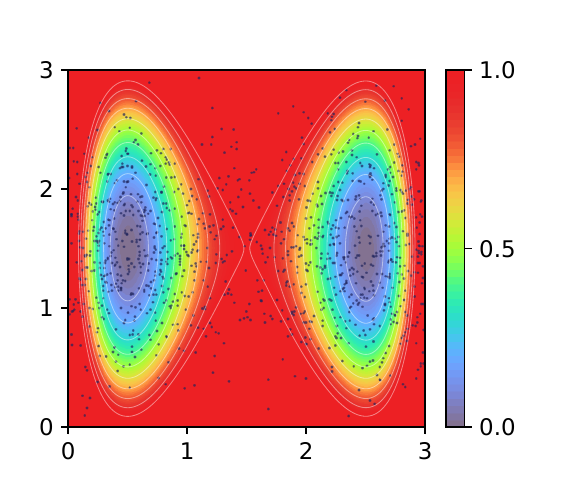}
\caption{Gaussian noise, $\sigma=0.5$.}
\end{subfigure}
\par\medskip
\begin{subfigure}[t]{0.32\textwidth}
\centering
\includegraphics[width=\linewidth]{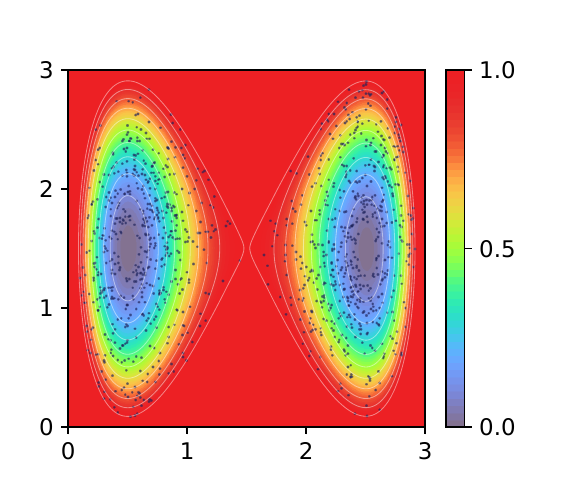}
\caption{Biased dynamics, $s=1$.}
\end{subfigure}
\hfill
\begin{subfigure}[t]{0.32\textwidth}
\centering
\includegraphics[width=\linewidth]{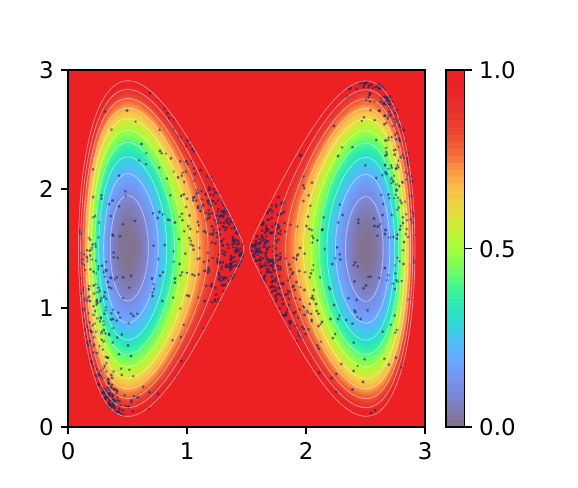}
\caption{Biased dynamics, $s=2$.}
\end{subfigure}
\hfill
\begin{subfigure}[t]{0.32\textwidth}
\centering
\includegraphics[width=\linewidth]{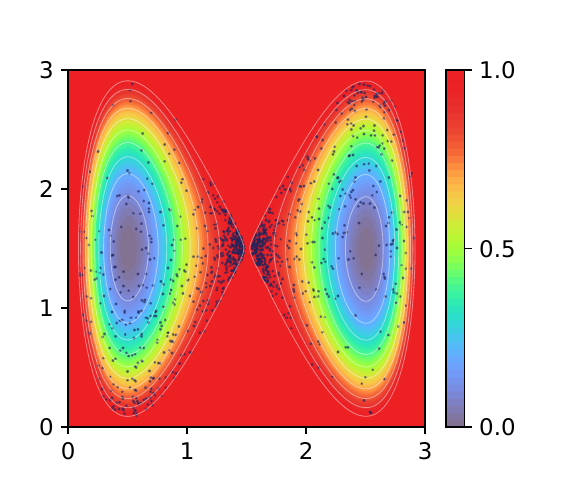}
\caption{Biased dynamics, $s=3$.}
\end{subfigure}
\caption{Selected HJB collocation diagnostics on the learned double-well
landscape ($D=0.05$). Dark dots are generated from the original diagnostic checkpoint. All
panels show the same learned potential, shifted to a minimum of zero,
with a common color scale and a separate colorbar in each panel.
(a--c) Increasing Gaussian perturbations broaden the flow samples. These
noise diagnostics redraw points that leave the domain, whereas the main
method uses reflection. (d--f) Increasing bias strength at fixed
$\Delta t=0.05$, with 50 steps and energy-window filtering on $[0.5D,20D]$.
The two comparisons vary noise strength and bias strength separately.}
\label{fig:enhancement}
\end{figure}

This subsection records the numerical settings and configuration-selection
procedure. Metrics are computed in float64 at the final epoch.
\Cref{fig:enhancement} shows the two HJB collocation mechanisms on the
learned double-well landscape.

\paragraph{Implications for collocation coverage}
\Cref{fig:enhancement}(a--c) shows that direct flow samples concentrate in
the two low-energy wells, leaving the interwell barrier sparsely sampled.
Weak Gaussian perturbations mainly broaden the sampled neighborhoods of the
wells; larger perturbations also populate the barrier and higher-energy
regions. In panels (d--f), increasing the bias strength directs a larger
fraction of the retained collocation points toward the barrier. Both
mechanisms therefore extend the spatial coverage of the HJB constraint
beyond the regions emphasized by the on-policy EPR measure.

Enhancement strength controls the tradeoff between barrier coverage and
the yield of accepted collocation points.
For the barrier band $|x_1-1.5|<0.3$, increasing $s$ from $2$ to $3$ raises
the fraction of retained points in this band from $30.8\%$ to $45.3\%$,
but lowers the fraction of candidate snapshots accepted by the energy window
from $13.9\%$ to $7.2\%$. These fractions are computed from the full sampled
pools; each panel displays the same number of retained points. Thus the
choice of enhancement strength should account for both spatial coverage and
the yield of admissible collocation points. This fixed-model diagnostic
characterizes the spatial redistribution of HJB points; the training
ablations in \ref{app:ablations} assess reconstruction accuracy.

\paragraph{Selection rule}
For the low-dimensional benchmarks and Gaussian mixture, candidate
configurations were ranked by the mean rRMSE.
Factors not under comparison were held fixed. The reported statistics
include the runs used for configuration selection.

\paragraph{Benchmark and baseline details}
The double-well and multi-stable domains are $[0,3]^2$; the limit-cycle
domain is $[0,8]\times[0,6]$. The twelve-dimensional mixture has component
weights $0.75$ and $0.25$, separated means, and equal covariances $0.03I$.
Its drift is $F=-D(J-I)\nabla\log p_{\mathrm{mix}}$, where $J$ is the
antisymmetric part of a fixed random matrix. The KL baseline trains
the same flow on $10^5$ stationary samples without either physics loss.

\paragraph{Flow}
Low-dimensional and biological-network problems use a bounded Bernstein
flow with $M=48$, $K=4$, conditioner width $96$, and $\eta=10^{-10}$.
The twelve-dimensional mixture uses the unbounded variant with $M=32$ and
$\eta=10^{-8}$.  Inversion uses $48$ bisection iterations, and a sample is
accepted in training only when the forward closure error is below
$10^{-3}$ in float32.

\paragraph{Shared training choices}
Training starts at $D_0=10D$. After a hold of $H$ epochs, the diffusion is
annealed over $A$ epochs according to
\begin{equation}
 D_k=D_0\left(\frac{D}{D_0}\right)^{g(f_k)},\qquad
 f_k=\frac{k-H}{A}\in[0,1],\qquad
 g(f)=3f^2-2f^3+3f^2(1-f)^2.
 \label{eq:anneal}
\end{equation}
The diffusion is held at $D$ after annealing. The interpolation has zero
slope at both endpoints; the learning rate follows a separate schedule.
For the benchmark configurations in \cref{tab:protocol}, the EPR pool
contains $10^4$ on-policy samples and is redrawn every $R$
epochs.  The diffusion schedule is \eqref{eq:anneal}.  Runs of
$50\,001$ epochs hold $H=1000$ epochs at $10D$ and anneal for
$A=2\times10^4$ epochs; the learning rate is held for $25\,000$ epochs and
then decayed geometrically to one tenth at the final epoch.
Unless otherwise specified in \cref{tab:protocol}, biased dynamics uses
$4096$ walkers, $200$ steps of size $0.05$ ($40$ burn-in, stride $20$),
$s=\eta=1$, energy window $[0.5D_k,25D_k]$, and a buffer of $2\times10^5$
points. Walker snapshots are accepted only if their energy relative to the
minimum in the fresh flow sample batch lies within the prescribed window;
during the initial hold, HJB points are instead noise-inflated
with $\sigma=0.5$.  Protocol selection compared $\lambda\in\{0.1,1,10\}$, together with
EPR-only ($\lambda=0$) and HJB-only training. Here $\lambda$ is the ratio
of the HJB and EPR coefficients in the archived objectives; the original
overall loss scaling is retained in the archived training configurations.

\begin{table}[t]
\centering
\caption{Final training protocol per problem.  Shared choices are stated
in the text of this subsection; the last column lists system-specific collocation settings.}
\label{tab:protocol}
\small
\begin{tabular}{lcccccc}
\toprule
System & Epochs & Batch & $R$ & $\lambda$ & lr$_0$ & HJB collocation \\
\midrule
Double well, $D=0.1$, EPR only & $50\,001$ & 4096 & $10$ & $0$ & $10^{-3}$ & --- \\
Double well, $D=0.05$ & $50\,001$ & 4096 & $10$ & $10$ & $10^{-3}$ & biased, refresh $10^3$ \\
Limit cycle, $D=0.1$ & $50\,001$ & 4096 & $10$ & $10$ & $10^{-3}$ & noise $\sigma=0.05$ \\
Multi-stable, $D=0.01$ & $50\,001$ & 4096 & $10$ & $10$ & $10^{-3}$ & biased, refresh $10^2$ \\
12-D mixture, $D=0.1$ & $50\,001$ & 4096 & $100$ & $0.1$ & $10^{-4}$ & noise $\sigma=0.25$ \\
\bottomrule
\end{tabular}
\end{table}

A noise-inflation variant of the multi-stable problem, with $\sigma=0.1$
and $\lambda=30$, is reported only as a secondary result,
with a mean rRMSE of $0.0320$ over three runs. The full-dimensional network experiments
and the reduced-force learning pipeline use the separate protocols in
\ref{app:dimension-reduction}.

\paragraph{Weak-form comparison}
Both WGS variants use $8000$ samples and $300$ test centers per update,
with centers jittered by $\varepsilon=0.6$, kernel width $\sigma=0.1$, and
Adam at learning rate $10^{-4}$ with weight decay $5\times10^{-5}$.  The
official generator is a float64 RealNVP of width $32$ and depth $3$.
Landscape rRMSE and rMAE are computed on the reference grid masked at $20D$.
The additional RealNVP configurations vary kernel width, kernel schedule, center count, jitter, and learning rate.

\subsection{Full-dimensional networks and dimensionality reduction}
\label{app:dimension-reduction}

\paragraph{Full-dimensional learning}
The cell-cycle and gene-network flows are trained on $[0,1]^3$ at $D=0.01$
and $[0,3]^{52}$ at $D=0.02$, respectively. Both use the bounded Bernstein
architecture with four blocks, $48$ components, width $96$, and
$\eta=10^{-10}$. The original analytic drift supplies all force evaluations.
The full-dimensional objective uses $\lambda=10$, with the EPR
squared norm summed over coordinates. Fresh model samples supply EPR;
Gaussian perturbations supply HJB, with no stationary sample pool or
high-temperature replay. The 52D implementation batches conditioner
computations and uses exact model scores and Laplacians. Training is in
float32 and checkpoint evaluation in float64.

\begin{table}[H]
\centering
\caption{Biological-network training stages. $R$ is the interval for
refreshing the EPR sample pool. Force regression uses fixed samples
drawn from the trained full-dimensional model.}
\label{tab:network-protocol}
\small
\begin{tabular}{lrrrl}
\toprule
Stage & Epochs & Batch & $R$ & Training data or HJB source \\
\midrule
Full 3D flow & $50\,001$ & $4096$ & $10$ & HJB noise, box width/$30$ \\
Full 52D flow & $10\,001$ & $512$ & $10$ & HJB noise, box width/$30$ or $/120$ \\
Reduced force & $1000$ & $2048$ & --- & $10\,000$ full-flow samples \\
DR flow (2D) & $100\,001$ & $4096$ & $100$ & HJB biased dynamics \\
\bottomrule
\end{tabular}
\end{table}

Full 3D training holds $10D$ for $1000$ epochs and anneals for $20\,000$;
full 52D training holds for $250$ and anneals for $4000$, using
\eqref{eq:anneal}. Initial learning rates are $10^{-3}$, held for
$25\,000$ and $5000$ epochs, then geometrically reduced to $10^{-4}$ over
$25\,000$ and $5000$ epochs, respectively. EPR pools contain $10\,000$
and $10\,240$ points; 52D training uses $2048$ points per epoch. HJB
warm-up uses a perturbation of box width/$6$ for the first $2000$ epochs
in 3D, and box width/$30$ for the first $500$ epochs in 52D.

\paragraph{Model-sample force regression and DR}
Retained coordinates are CDK1--Plk1 and GATA6--NANOG. For each full model,
$10\,000$ direct samples and their known projected forces define
\eqref{eq:model-force-loss}. The force network has two inputs, two outputs,
and three tanh hidden layers of width $80$; Adam uses learning rate
$10^{-3}$. The resulting force is frozen for all reduced fits.
The 2D flow has the same Bernstein architecture. In this implementation,
the EPR residual is averaged over its two coordinates: the configured
EPR coefficient $0.1$, with unit HJB coefficient, therefore corresponds to
$\lambda=20$ in the squared-norm convention of the main text.
Its diffusion holds at $10D$ for $2000$ epochs and decreases geometrically
to $D$ over $58\,000$ epochs (linear exponent, rather than the shaped
exponent of \eqref{eq:anneal}). The learning rate is $10^{-3}$ for
$31\,000$ epochs, then decays to $10^{-4}$ over $29\,000$ epochs.
The EPR pool contains $10\,000$ fresh model samples. For both networks,
HJB uses $4096$ biased walkers, $50$ steps of size $10^{-3}$, burn-in $10$,
stride $5$, and $s=\eta=1$. The energy window is $[0.5D_k,20D_k]$, with
buffer size $200\,000$ refreshed every $1000$ epochs; the first $2000$
epochs use Gaussian perturbations of box width/$30$ instead. There is no
high-temperature replay.

\paragraph{Sensitivity to full-model collocation}
For the 52D system, full-dimensional HJB perturbations at box width/$30$
give a mean DR correlation of $0.9143$ and a mean aligned rRMSE of $0.2850$.
At box width/$120$, the corresponding values are $0.9245$ and $0.2696$.
Both configurations recover the same principal landscape structure.

The reported DR metrics were recomputed from the archived final
checkpoints on the packaged reference grids. Correlation and aligned rRMSE use reference cells below $30D$.
The reported means average three reduced fits sharing the same full model
and regressed force. The standard deviations of correlation and aligned
rRMSE are $0.0005$ and $0.0008$ for the 3D system. For the 52D system,
they are $0.0006$ and $0.0003$ at box width/$30$, and $0.0002$ and
$0.0006$ at box width/$120$. Figure~\ref{fig:dr-landscapes}
uses the saved $10\,000$ full-flow samples per system, with a fixed subset
for overlays, and evaluates streamlines from the archived force networks.
The corresponding plotting source and checkpoint checksums accompany the
figure. These comparisons assess the learned two-coordinate stationary
marginal landscapes.

\subsection{Ablation studies}
\label{app:ablations}

To assess the contribution of each component, the study compares, one factor at
a time:
\begin{enumerate}[leftmargin=*,itemsep=2pt]
  \item self-sampled EPR alone versus Enhanced EPR, and HJB alone;
  \item the HJB collocation mechanisms of \cref{sec:enhancement}: noise
        inflation at several noise levels, biased dynamics, and fresh
        on-policy points;
  \item the HJB weights $\lambda=0.1$, $1$, and $10$, together with HJB-only training;
  \item direct training at the target $D$ versus diffusion continuations of
        several anneal lengths; and
  \item the on-policy redraw interval, comparing $R=100$ with $R=10$.
\end{enumerate}
Each ablation reports the mean landscape error and its standard deviation.
\Cref{tab:ablation-p1}
summarizes the single-factor double-well experiments used to select the
reference protocol.

\begin{table}[t]
\centering
\caption{Single-factor ablations on the double well at $D=0.05$.
Results are reported as means with standard deviations.
The reference setting is
noise inflation $\sigma=0.5$, $\lambda=1$, anneal
$A=4\times10^4$, $R=100$; each block varies one factor.}
\label{tab:ablation-p1}
\footnotesize
\begin{minipage}[t]{0.48\linewidth}
\centering
\begin{tabular}{lc}
\toprule
Setting & rRMSE \\
\midrule
\multicolumn{2}{l}{\emph{HJB collocation ($\lambda=1$, $A=4\times10^4$)}} \\
fresh on-policy, $\sigma=0$ & \withsd{0.3125}{0.3094} \\
noise $\sigma=0.05$ & \withsd{0.1322}{0.1186} \\
noise $\sigma=0.1$ & \withsd{0.0576}{0.0159} \\
noise $\sigma=0.25$ & \withsd{0.0218}{0.0104} \\
noise $\sigma=0.5$ & \withsd{0.0067}{0.0018} \\
biased dynamics & \withsd{0.0067}{0.0020} \\
\midrule
\multicolumn{2}{l}{\emph{HJB weight $\lambda$ ($A=4\times10^4$)}} \\
$\lambda=0.1$, noise $\sigma=0.5$ & \withsd{0.0520}{0.0304} \\
$\lambda=0.1$, biased & \withsd{0.0558}{0.0369} \\
$\lambda=1$, noise $\sigma=0.5$ & \withsd{0.0067}{0.0018} \\
$\lambda=1$, biased & \withsd{0.0067}{0.0020} \\
$\lambda=10$, noise $\sigma=0.5$ & \withsd{0.0036}{0.0011} \\
$\lambda=10$, biased & \withsd{0.0036}{0.0013} \\
HJB only, $\sigma=0.5$ & \withsd{0.0147}{0.0084} \\
HJB only, biased & \withsd{0.0031}{0.0042} \\
\bottomrule
\end{tabular}
\end{minipage}\hfill
\begin{minipage}[t]{0.48\linewidth}
\centering
\begin{tabular}{lc}
\toprule
Setting & rRMSE \\
\midrule
\multicolumn{2}{l}{\emph{Anneal length $A$ ($\lambda=10$)}} \\
none, noise $\sigma=0.5$ & \withsd{0.3111}{0.2721} \\
none, biased & \withsd{0.4724}{0.1691} \\
$10^4$, noise $\sigma=0.5$ & \withsd{0.0089}{0.0088} \\
$10^4$, biased & \withsd{0.0032}{0.0012} \\
$2\times10^4$, noise $\sigma=0.5$ & \withsd{0.0028}{0.0007} \\
$2\times10^4$, biased & \withsd{0.0034}{0.0011} \\
$4\times10^4$, noise $\sigma=0.5$ & \withsd{0.0036}{0.0011} \\
$4\times10^4$, biased & \withsd{0.0036}{0.0013} \\
\midrule
\multicolumn{2}{l}{\emph{Redraw $R$ ($\lambda=10$, $A=2\times10^4$)}} \\
$R=100$, noise $\sigma=0.5$ & \withsd{0.0028}{0.0007} \\
$R=10$, noise $\sigma=0.5$ & \withsd{0.0020}{0.0002} \\
$R=100$, biased & \withsd{0.0034}{0.0011} \\
$R=10$, biased & \withsd{0.0010}{0.0005} \\
\bottomrule
\end{tabular}
\end{minipage}
\end{table}

Three trends emerge within this double-well study. First, enhanced HJB
collocation lowers the error relative to unperturbed flow samples, with
larger noise amplitudes performing better over the tested range. Second,
removing diffusion continuation substantially increases the mean error for
both collocation mechanisms; the effect of anneal length depends on the
mechanism. Third, the tested
positive HJB weights favor $\lambda=10$, and more frequent pool refreshes
further improve the biased-dynamics result. HJB-only training also performs
well here. These findings guide the choice of the double-well training
protocol.

\clearpage
\clearpage

\begin{thebibliography}{10}
\expandafter\ifx\csname url\endcsname\relax
  \def\url#1{\texttt{#1}}\fi
\expandafter\ifx\csname urlprefix\endcsname\relax\def\urlprefix{URL }\fi
\expandafter\ifx\csname href\endcsname\relax
  \def\href#1#2{#2} \def\path#1{#1}\fi

\bibitem{Waddington57}
C.~H. Waddington, The Strategy of the Genes, George Allen \& Unwin, Ltd.,
  London, 1957.

\bibitem{Wang15Review}
J.~Wang, Landscape and flux theory of non-equilibrium dynamical systems with
  application to biology, Adv. Phys. 64~(1), 1--137, 2015.

\bibitem{Fang19Review}
X.~Fang, K.~Kruse, T.~Lu, J.~Wang, Nonequilibrium physics in biology, Rev. Mod.
  Phys. 91~(4), 045004, 2019.

\bibitem{Wang10}
J.~Wang, C.~Li, E.~Wang, Potential and flux landscapes quantify the stability
  and robustness of budding yeast cell cycle network, Proc. Natl. Acad. Sci.
  U.S.A. 107~(18), 8195--8200, 2010.

\bibitem{Lv14}
C.~Lv, X.~Li, F.~Li, T.~Li, Constructing the energy landscape for genetic
  switching system driven by intrinsic noise, PLoS {ONE} 9~(2), e88167, 2014.

\bibitem{Li16}
C.~Li, T.~Hong, Q.~Nie, Quantifying the landscape and kinetic paths for
  epithelial--mesenchymal transition from a core circuit, Phys. Chem. Chem.
  Phys. 18~(27), 17949--17956, 2016.

\bibitem{Shi22}
J.~Shi, K.~Aihara, T.~Li, L.~Chen, Energy landscape decomposition for cell
  differentiation with proliferation effect, Natl. Sci. Rev. 9~(8), nwac116,
  2022.

\bibitem{zhao2024nsr}
Y.~Zhao, W.~Zhang, T.~Li, {EPR-Net}: constructing a non-equilibrium potential
  landscape via a variational force projection formulation, Natl. Sci. Rev.
  11~(7), nwae052, 2024.

\bibitem{Zhou16}
P.~Zhou, T.~Li, Construction of the landscape for multi-stable systems:
  Potential landscape, quasi-potential, {A}-type integral and beyond, J. Chem.
  Phys. 144~(9), 094109, 2016.

\bibitem{Cameron12}
M.~K. Cameron, Finding the quasipotential for nongradient {SDE}s, Physica D
  241~(18), 1532--1550, 2012.

\bibitem{Ge12}
H.~Ge, H.~Qian, Landscapes of non-gradient dynamics without detailed balance:
  Stable limit cycles and multiple attractors, Chaos 22~(2), 023140, 2012.

\bibitem{Zhou12}
J.~X. Zhou, M.~D.~S. Aliyu, E.~Aurell, S.~Huang, Quasi-potential landscape in
  complex multi-stable systems, J. R. Soc. Interface 9~(77), 3539--3553, 2012.

\bibitem{Maier93}
R.~S. Maier, D.~L. Stein, Escape problem for irreversible systems, Phys. Rev. E
  48~(2), 931--938, 1993.

\bibitem{Aurell02}
E.~Aurell, K.~Sneppen, Epigenetics as a first exit problem, Phys. Rev. Lett.
  88~(4), 048101, 2002.

\bibitem{Zhang12}
X.-J. Zhang, H.~Qian, M.~Qian, Stochastic theory of nonequilibrium steady
  states and its applications. part i, Phys. Rep. 510~(1--2), 1--86, 2012.

\bibitem{Lin21}
B.~Lin, Q.~Li, W.~Ren, A data driven method for computing quasipotentials, in:
  Proceedings of the 2nd Mathematical and Scientific Machine Learning
  Conference, Vol. 145 of Proceedings of Machine Learning Research, PMLR, pp.
  652--670, 2022.

\bibitem{Lin22}
B.~Lin, Q.~Li, W.~Ren, Computing the invariant distribution of randomly
  perturbed dynamical systems using deep learning, J. Sci. Comput. 91~(3), 77,
  2022.

\bibitem{Lin23}
B.~Lin, Q.~Li, W.~Ren, Computing high-dimensional invariant distributions from
  noisy data, J. Comput. Phys. 474, 111783, 2023.

\bibitem{Raissi19}
M.~Raissi, P.~Perdikaris, G.~E. Karniadakis, Physics-informed neural networks:
  A deep learning framework for solving forward and inverse problems involving
  nonlinear partial differential equations, J. Comput. Phys. 378, 686--707,
  2019.

\bibitem{rezende2015variational}
D.~J. Rezende, S.~Mohamed, Variational inference with normalizing flows, in:
  Proceedings of the 32nd International Conference on Machine Learning, Vol.~37
  of Proceedings of Machine Learning Research, PMLR, pp. 1530--1538, 2015.

\bibitem{papamakarios2021normalizing}
G.~Papamakarios, E.~Nalisnick, D.~J. Rezende, S.~Mohamed, B.~Lakshminarayanan,
  Normalizing flows for probabilistic modeling and inference, J. Mach. Learn.
  Res. 22~(57), 1--64, 2021.

\bibitem{tang2022adda}
K.~Tang, X.~Wan, Q.~Liao, Adaptive deep density approximation for
  {Fokker--Planck} equations, J. Comput. Phys. 457, 111080, 2022.

\bibitem{zeng2025boundedkrnet}
L.~Zeng, X.~Wan, T.~Zhou, Bounded {KRnet} and its applications to density
  estimation and approximation, SIAM J. Sci. Comput. 47~(6), C1294--C1318,
  2025.

\bibitem{Zhang16}
W.~Zhang, C.~Hartmann, C.~Sch{\"u}tte, Effective dynamics along given reaction
  coordinates{,} and reaction rate theory, Faraday Discuss. 195, 365--394,
  2016.

\bibitem{liu2026fesfm}
Z.~Liu, T.~Li, {FES-FM}: Free energy surface sampling via reduced flow
  matching, arXiv preprint arXiv:2605.00337, 2026.

\bibitem{cai2026wgs}
Z.~Cai, Y.~Cao, Y.~Huang, X.~Zhou, Weak generative sampler to efficiently
  sample invariant distribution of stochastic differential equation, SIAM J.
  Sci. Comput. 48~(4), C708--C735, 2026.

\bibitem{Wang08}
J.~Wang, L.~Xu, E.~Wang, Potential landscape and flux framework of
  nonequilibrium networks: Robustness, dissipation, and coherence of
  biochemical oscillations, Proc. Natl. Acad. Sci. U.S.A. 105~(34),
  12271--12276, 2008.

\bibitem{Wang11}
J.~Wang, K.~Zhang, L.~Xu, E.~Wang, Quantifying the {W}addington landscape and
  biological paths for development and differentiation, Proc. Natl. Acad. Sci.
  U.S.A. 108~(20), 8257--8262, 2011.

\bibitem{Ferrell11}
J.~E. Ferrell, Jr., T.~Y.-C. Tsai, Q.~Yang, Modeling the cell cycle: Why do
  certain circuits oscillate?, Cell 144~(6), 874--885, 2011.

\bibitem{Li13}
C.~Li, J.~Wang, Quantifying cell fate decisions for differentiation and
  reprogramming of a human stem cell network: Landscape and biological paths,
  PLoS Comput. Biol. 9~(8), e1003165, 2013.

\bibitem{liu2025scoresingularity}
Z.~Liu, W.~Zhang, T.~Li, Improving the {Euclidean} diffusion generation of
  manifold data by mitigating score function singularity, in: Advances in
  Neural Information Processing Systems, Vol.~38, 2025.

\bibitem{liu2026rddpm}
Z.~Liu, W.~Zhang, C.~Sch{\"u}tte, T.~Li, {Riemannian} denoising diffusion
  probabilistic models, Commun. Math. Sci. 24~(5), 1267--1295, 2026.

\bibitem{hyvarinen05a}
A.~Hyv{\"a}rinen, Estimation of non-normalized statistical models by score
  matching, J. Mach. Learn. Res. 6~(24), 695--709, 2005.

\bibitem{Song2019}
Y.~Song, S.~Ermon, Generative modeling by estimating gradients of the data
  distribution, in: Advances in Neural Information Processing Systems, Vol.~32,
  Curran Associates, Inc., 2019.

\bibitem{nardi2014neumann}
G.~Nardi, Schauder estimate for solutions of {Poisson}'s equation with
  {Neumann} boundary condition, Enseign. Math. 60~(3--4), 421--435, 2014.

\bibitem{ramasinghe2022bernstein}
S.~Ramasinghe, K.~Fernando, S.~Khan, N.~Barnes, A robust normalizing flow using
  {B}ernstein-type polynomials, in: 33rd British Machine Vision Conference,
  BMVA Press, 2022.

\bibitem{durr2024bernstein}
O.~D{\"u}rr, S.~H{\"o}rtling, D.~Dold, I.~Kovylov, B.~Sick, Bernstein flows for
  flexible posteriors in variational {B}ayes, AStA Adv. Stat. Anal. 108~(2),
  375--394, 2024.

\bibitem{amorese2026bernstein}
P.~Amorese, M.~Lahijanian, Universal learning of stochastic dynamics for exact
  belief propagation using {B}ernstein normalizing flows, in: Proceedings of
  the {AAAI} Conference on Artificial Intelligence, Vol.~40, pp. 36610--36617,
  2026.

\end{thebibliography}
\end{document}